\documentclass{article} 
\usepackage{iclr2027_conference,times}

\usepackage{amsmath,amsfonts,bm}

\def\eqref#1{equation~\ref{#1}}

\def\1{\bm{1}}

\DeclareMathAlphabet{\mathsfit}{\encodingdefault}{\sfdefault}{m}{sl}
\SetMathAlphabet{\mathsfit}{bold}{\encodingdefault}{\sfdefault}{bx}{n}

\usepackage[T1]{fontenc}
\usepackage{graphicx}
\usepackage{amssymb,amsthm}
\usepackage{booktabs}
\usepackage{xcolor,colortbl}
\usepackage{subcaption}
\usepackage{enumitem}
\usepackage{wrapfig}
\usepackage{caption}

\newtheorem{theorem}{Theorem}[section]

\newtheorem{proposition}[theorem]{Proposition}

\usepackage{hyperref}
\usepackage{url}

\renewcommand{\eqref}[1]{\textup{(\ref{#1})}}

\title{CoMNO: Commutator Memory Neural Operators for Long-Horizon PDE Prediction}

\author{%
{\bfseries Han Xiao$^{\dagger}$, Xin Zhang$^{\dagger}$, Yuhan Zou} \\
{\bfseries Zhenzhong Wang$^{*}$, Donglin Cao$^{*}$, Dazhen Lin$^{*}$, Min Jiang$^{*}$} \\
{\normalfont Xiamen University} \\
{\small\normalfont $^{\dagger}$Equal contribution} \\
{\small\normalfont
\texttt{zhenzhongwang@xmu.edu.cn},
\texttt{another@xmu.edu.cn}} \\
{\small\normalfont
\texttt{dzlin@xmu.edu.cn},
\texttt{minjiang@xmu.edu.cn}}%
}

\iclrfinalcopy
\hypersetup{hidelinks}

\makeatletter
\renewcommand{\paragraph}{\@startsection{paragraph}{4}{\z@}{0.8ex}{-1em}{\normalsize\bf}}
\makeatother

\begin{document}
\raggedbottom
\setlength{\parskip}{4pt}

\maketitle
\pagestyle{fancy}
\thispagestyle{fancy}
\lhead{\strut}

\begin{abstract}
Solving nonlinear partial differential equations (PDEs) over long time horizons is computationally expensive using traditional numerical methods. While neural operators provide an efficient alternative, they suffer from the accumulation of approximation errors along the predicted trajectory, typically caused by the failure to capture fine-scale details.
In this work, we design a Commutator Memory Neural Operator that identifies fine-scale dynamical details by commuting two operations and leverages these details to enhance the neural operator's capability in capturing fine-scale features over long time horizons. Specifically, we identify a performance gap between two operations: downsampling the high-resolution field before rollout prediction versus performing rollout prediction before downsampling. We observe that fine-scale dynamical details are often encoded in this gap. By commuting these two operations, the fine-scale details removed during downsampling can be captured. Furthermore, we employ a memory module to store these fine-scale features along the trajectory and retrieve them during rollout, generating spatially adaptive corrections that enhance long-horizon fidelity. Empirical results across five systems demonstrate that CoMNO improves prediction accuracy and long-horizon rollout stability.
\end{abstract}

\begin{wrapfigure}{r}{0.50\textwidth}
    \centering
    \vspace{-8pt}
    \includegraphics[
        width=0.98\linewidth,
        keepaspectratio
    ]{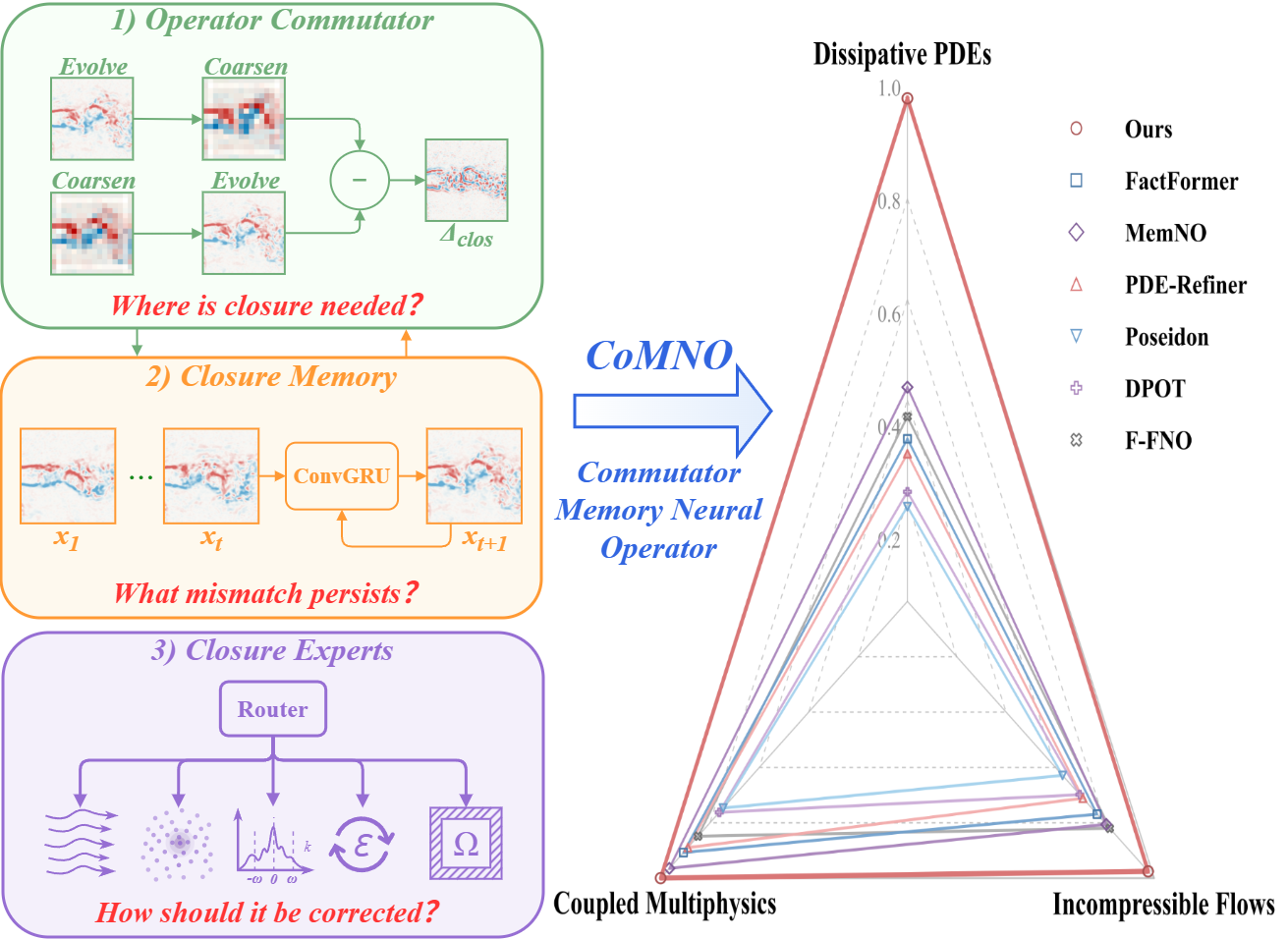}
    \caption{CoMNO's core mechanisms and gains in long-horizon prediction across diverse PDE systems.}
    \label{fig:intro}
    \vspace{-10pt}
\end{wrapfigure}

\section{Introduction}

High-fidelity simulation of nonlinear partial differential equations (PDEs) is central to climate modeling, turbulence analysis, and engineering design~\cite{lam2023graphcast,kochkov2021cfd,pfaff2021meshgraphnets}.
Conventional numerical solvers require fine grids and repeated time integration, making long-horizon simulation computationally expensive. To address this efficiency bottleneck, neural operators offer an efficient alternative by learning mappings between physical fields~\cite{li2021fno,kovachki2023neuraloperator}.

Despite remarkable progress, neural operators typically suffer from performance deterioration during long-time rollout prediction, due to the accumulation of approximation errors along the predicted trajectory ~\cite{lippe2023pderefiner,koehler2024apebench,wang2026cross}. 
This difficulty can be viewed as a finite-resolution closure problem~\cite{gupta2021neural,ICLR2025_89379d5f}. To formalize this, let \(u_t\) denote the high-resolution field, 
\(\Phi\) the neural operator that maps it forward in time, and 
\(D_0\) a downsampling operator that extracts the coarse observation 
\(x_t = D_0 u_t\). Consider the "predict-then-downsample" 
pipeline: the neural operator first predicts the next high-resolution 
state \(\Phi(u_t)\), which is then downsampled to yield 
\(D_0\Phi(u_t)\). This composite step can be expressed as:
\begin{equation}
x_{t+1}
=
D_0\Phi(u_t)
=
x_t+\Delta_{\mathrm{res}}(x_t)
+\mathcal{R}(x_t,\mathcal{H}_t),
\label{eq:closure_formulation}
\end{equation}
where the right-hand side is interpreted as an update from the current coarse state \(x_t\). The term $\Delta_{\mathrm{res}}(x_t)$ denotes the one-step state increment from $x_t$ predicted solely from the current coarse-scale state, while the term \(\mathcal{R}(x_t,\mathcal{H}_t)\) captures the residual influence of fine-scale structures that were discarded by downsampling but continue to affect the future coarse evolution. This residual correction is necessary because the coarse state \(x_t\) does not uniquely determine the future. The history \(\mathcal{H}_t\) thus supplies the missing information needed to infer this fine-scale influence.

The above formulation suggests two key requirements for addressing this problem: first, one must identify whether the fine-scale information lost during downsampling actually matters for the future coarse rollout prediction; second, if it does, one must retain and utilize this information over time to enhance prediction fidelity. Existing approaches tackle this challenge from various angles. Multiscale methods refine spatial representation to preserve finer structures, memory-based models augment temporal modeling to capture long-range dependencies, and iterative correction schemes refine predictions through repeated feedback~\cite{lippe2023pderefiner,ICLR2025_89379d5f}. However, these methods typically infer missing dynamics from coarse states or latent features without explicitly tracing what information disappears during downsampling and how its removal changes future rollout predictions.

In this work, we find that the fine-scale influence underlying
\(\mathcal{R}(x_t,\mathcal{H}_t)\) can be identified by commuting rollout
prediction and downsampling. For each scale \(s\), let \(D_s\) denote the
downsampling operator, and apply the same neural operator \(\Phi\) in both
paths. We define the commutator residual at time \(t\) and scale \(s\) as
\begin{equation}
c_t^{(s)}
=
D_s\Phi(u_t)
-
\Phi(D_su_t),
\label{eq:commutator}
\end{equation}
The first path evolves \(u_t\) before removing scale \(s\), whereas the
second removes that scale before evolution. Their discrepancy therefore
measures how the discarded fine-scale structures alter the predicted coarse
state, providing direct evidence for their residual contribution to future
coarse dynamics.


To compensate for fine-scale effects lost during coarse-resolution rollout,
we introduce the Commutator Memory Neural Operator (CoMNO). Its key
idea is that commutator residuals reveal where unresolved structures alter
the predicted coarse evolution and how strongly they do so. A single residual
captures this influence only at the current step, although its effect may
persist and reappear later. CoMNO therefore stores residual evidence along the
trajectory, integrates it across scales and time, and uses the accumulated
history to determine where and how strongly future predictions should be
corrected. In this way, CoMNO turns transient cross-scale discrepancies into
spatially adaptive corrections, reducing the repeated loss of unresolved
dynamics and stabilizing long-horizon rollout.

Our contributions are summarized as follows:
\begin{enumerate} 
    \item We identify fine-scale dynamical details through the prediction gap
    between downsampling before rollout prediction and performing rollout
    prediction before downsampling.

    \item We introduce CoMNO, which stores these details in recurrent Closure
    Memory and retrieves them to generate spatially adaptive corrections
    throughout autoregressive rollout.

    \item Across three APEBench systems and two RealPDEBench systems
    \cite{koehler2024apebench,hu2026realpdebench}, CoMNO consistently improves
    long-horizon accuracy, rollout stability, spectral fidelity, and physical
    consistency over strong neural PDE baselines.
\end{enumerate}

\section{Related Work}

\paragraph{Neural Operators and Long-Horizon PDE Prediction.}
Neural operators learn mappings between function spaces using integral
kernels, spectral transforms, multiscale convolutions, attention, and graph
or coordinate representations
\cite{lu2021deeponet,kovachki2023neuraloperator,li2021fno,
tran2023factorized,rahman2023uno,raonic2023cno,
cao2021galerkin,li2023factformer,hao2024dpot,herde2024poseidon,
hao2023gnot,serrano2023coral,pfaff2021meshgraphnets,
brandstetter2022mppde}.
Despite strong short-horizon accuracy
\cite{takamoto2022pdebench}, autoregressive rollout accumulates distribution
shift, phase error, and spectral distortion
\cite{list2025unrolling,lippe2023pderefiner,
kohl2024benchmarkingautoregressiveconditionaldiffusion}.
Multi-step training, physics-informed objectives, refinement, and
probabilistic prediction improve stability
\cite{list2025unrolling,li2024pino,lippe2023pderefiner}, yet do not localize
unresolved corrections. CoMNO derives an online spatial signal for   this purpose.

\paragraph{Closure Modeling and Operator Non-Commutation.}
Large-eddy simulation represents unresolved scales through subgrid closures
\cite{zhao2023sgsassessment,sun2024dynamicsgs,
johnson2022physicsinspiredles}, while data-driven methods learn subgrid
stresses, fluxes, or coarse-grid corrections
\cite{maulik2019subgrid,beck2019les,kochkov2021cfd,
srinivasan2024localclosure,vangastelen2024energyclosure,
agdestein2025closure}.
Coarsening and nonlinear evolution generally do not commute, yielding
state-dependent residuals across space and resolution
\cite{sanderse2025closure,vangastelen2024energyclosure,
agdestein2025closure,yalla2021resolution}, and coarse chaotic states may
admit nonunique deterministic closures
\cite{wang2026coarsegrainingneuraloperators}.
Prior work mainly uses non-commutation for analysis or supervised closure
learning. CoMNO evaluates it online to allocate corrections where
cross-resolution inconsistency appears.

\paragraph{Non-Markovian Reduction and Memory-Augmented Dynamics.}
The Mori-Zwanzig formalism decomposes projected dynamics into Markovian,
history-dependent, and unresolved fluctuation terms
\cite{gilani2021nonmarkovian,lin2021koopmanmz,
falkena2019delaymz}.
Finite-memory closures and reduced models capture persistent unresolved effects
\cite{parish2017dynamicmz,parish2017nonmarkovian,
lin2021datamz,tian2021isotropicmz,wang2020rnnclosure}, while recurrent and reservoir
models retain trajectory history
\cite{vlachas2018lstm,pathak2018reservoir,
vlachas2020forecasting}, including MemNO
\cite{ICLR2025_89379d5f}.
These methods update memory from states without measuring scale mismatch.
CoMNO retains commutator evidence in Closure Memory to guide spatially adaptive corrections through Closure Experts.

\section{Methodology}

\subsection{Problem Formulation and Overview}
\label{sec:framework}

\begin{figure*}[t]
    \centering
    \includegraphics[width=\textwidth]{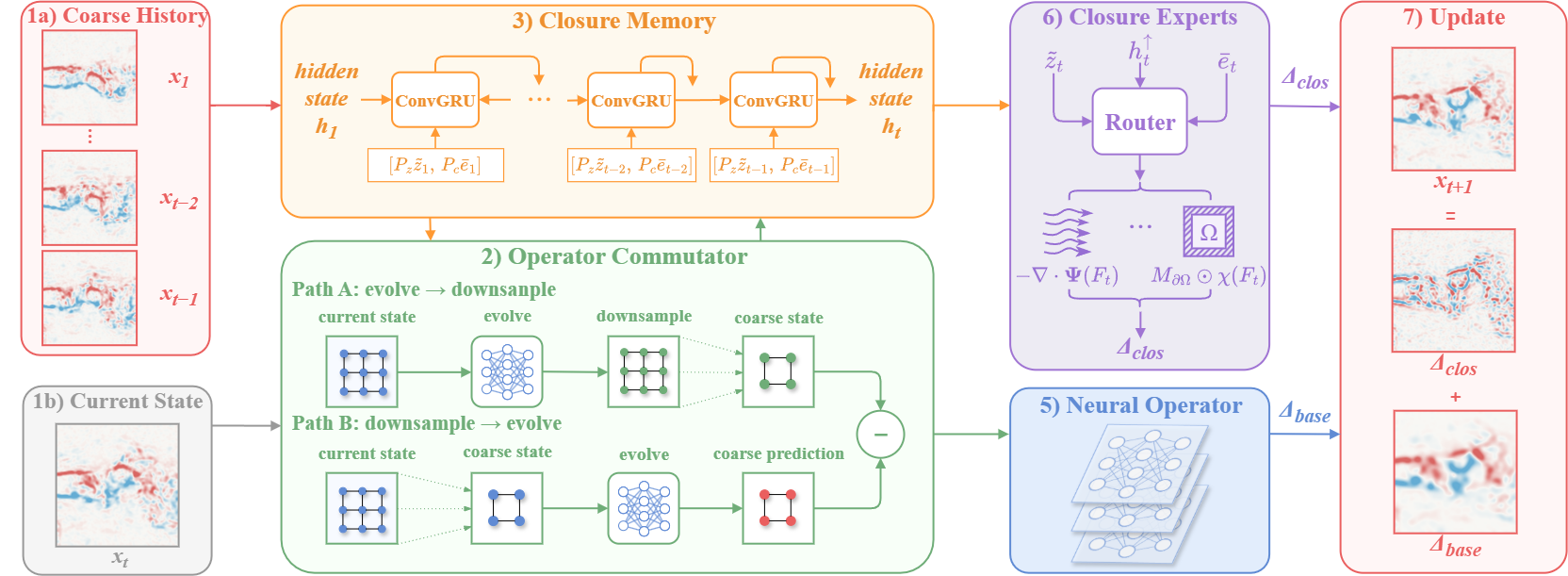}
    \caption{Overview of CoMNO: Operator Commutator evidence and Closure Memory jointly guide Closure Experts to produce spatially adaptive corrections.}
    \label{fig:method}
\end{figure*}

CoMNO learns the history-dependent closure in
Eq.~\eqref{eq:closure_formulation} by comparing predictions
across resolutions and retaining the resulting evidence
over time. Following the Introduction, the Current State
is the coarse observation \(x_t=D_0u_t\), and the Coarse
History provides the temporal context stored in \(h_t\).
Figure~\ref{fig:method} shows the computation: the Operator
Commutator measures how coarsening changes a prediction,
Closure Memory retains these differences along the
trajectory, and Closure Experts use this information
to correct the Neural Operator's base increment.

\subsection{Operator Commutator}
\label{sec:commutator}

The Operator Commutator measures where and how spatial
coarsening changes the forecast, providing evidence
for the closure correction.
It applies the comparison in Eq.~\eqref{eq:commutator}
at observable resolutions: the current field \(x_t\)
is compared with its downsampled version \(D_sx_t\).
Here, \(D_s\) performs spatial downsampling on the grid
of its input, and the same factor is applied to memory.

Both paths use the shared prediction map \(\Phi^0(x;h)\).
It advances the field using the Neural Operator's base
increment and the closure conditioned on memory.
The superscript \(0\) indicates that the current
commutator evidence is set to zero for this preliminary
prediction. We denote the online discrepancy by
\(\tilde{c}_t^{(s)}\) to distinguish it from the
high-resolution quantity in the Introduction:
\begin{equation}
\tilde{c}_t^{(s)}
=
\underbrace{D_s\Phi^0(x_t;h_t)}_{
\text{Path A: evolve, then downsample}}
-
\underbrace{\Phi^0(D_sx_t;D_sh_t)}_{
\text{Path B: downsample, then evolve}}.
\label{eq:online_commutator}
\end{equation}
Path A lets structures represented on the observation
grid affect the prediction before downsampling.
Path B removes these structures first, then predicts
from the coarser field and memory.
The two paths share parameters, advance over the same
time interval, and produce states on the same grid.
Subtracting their outputs measures how jointly coarsening
the field and memory changes the forecast.

The spatial pattern locates the affected regions.
For each state component, a positive discrepancy means
that Path A predicts a larger value than Path B at that
location. Its magnitude measures the size of this difference.
Repeating the comparison across scales tests how the
forecast changes as different amounts of spatial
information are removed.
CoMNO learns through rollout training how these
observable differences inform corrections on the
observation grid.

The same comparison is applied to the provisional
memory outputs to measure how coarsening changes the
information carried forward. A scale difference can
affect both the immediate prediction and the context
used for later predictions. Combining state and memory
discrepancies captures both effects in the evidence.

State and memory discrepancies are normalized by the
magnitudes of their respective Path A outputs.
This expresses each mismatch relative to its output
scale while preserving its sign and spatial variation.
The discrepancies are then lifted to the observation
grid, concatenated along channels, and averaged across
scales to form \(\bar{e}_t\).
Lifting aligns the evidence with the locations to be
corrected. Concatenation keeps state and memory
differences in separate channels.
The final prediction and memory update use this evidence
from the same starting pair \((x_t,h_t)\).

\subsection{Closure Memory}
\label{sec:memory}

Closure Memory retains commutator evidence so that
differences observed at earlier steps can guide later
corrections. A single commutator records the effect of
coarsening at the current step. Its history shows how
this effect develops as the field evolves, providing
context for interpreting each new discrepancy.

The Coarse History is processed in temporal order
to initialize this context.
Local features describe the current field and its
spatial derivatives, capturing local variation and
curvature. Conditioning them on \(h_t\) and
\(\bar{e}_t\) produces the features \(\tilde{z}_t\)
shown in Figure~\ref{fig:method}.

The ConvGRU updates memory as
\begin{equation}
h_{t+1}
=
\operatorname{ConvGRU}
\left([P_z\tilde{z}_t,P_c\bar{e}_t],h_t\right),
\label{eq:memory_update}
\end{equation}
where \(P_z\) and \(P_c\) project features and evidence
onto the memory grid, and brackets denote channel
concatenation.
The projected features describe the local structure,
the projected evidence records its current scale
discrepancy, and \(h_t\) supplies the preceding context.
Concatenation presents both current inputs to the gates,
which determine how to combine them with the existing
memory. Convolutional recurrence preserves the spatial
layout while updating this temporal information.

Retaining field features together with the evidence
allows memory to associate past discrepancies with
the structures that produced them. Similar current
fields can therefore receive different corrections
when their preceding trajectories differ.
The current correction uses \(h_t\) and \(\bar{e}_t\);
\(h_{t+1}\) carries the updated context to the next step.

\subsection{Neural Operator and Closure Experts}
\label{sec:closure}

The Neural Operator predicts the base increment
\(\Delta_{\mathrm{base}}(x_t)\) from the current field.
It implements \(\Delta_{\mathrm{res}}(x_t)\) in the
Introduction. Closure Experts provide the additional
correction needed to approximate the history-dependent
term \(\mathcal{R}(x_t,\mathcal{H}_t)\).
The Router and Closure Experts receive
\begin{equation}
F_t
=
[\tilde{z}_t,h_t^{\uparrow},\bar{e}_t],
\label{eq:closure_features}
\end{equation}
where \(h_t^{\uparrow}\) denotes memory lifted to the
observation grid. Concatenation aligns local field
features, past context, and current evidence at each
spatial location. The field features describe the
structure being corrected, memory describes its recent
evolution, and the evidence describes its sensitivity
to coarsening.
The Router uses these inputs to select expert
contributions. The experts use them to determine the
direction and magnitude of their corrections.

The Router selects how strongly each expert contributes.
It applies a softmax across experts at each location
and combines their outputs as
\begin{equation}
\Delta_{\mathrm{clos}}(\mathbf{r})
=
\sum_k\alpha_{t,k}(\mathbf{r})E_k(F_t)(\mathbf{r}),
\label{eq:closure_correction}
\end{equation}
where \(E_k\) is expert \(k\), \(\alpha_{t,k}\) is its
routing weight, and \(\mathbf{r}\) is a spatial location.
The weights are nonnegative and sum to one across
experts. Each expert supplies a candidate correction,
and its weight controls its contribution to the final
sum. Spatially varying weights let different regions
use different expert combinations within the same
prediction step.

Figure~\ref{fig:method} shows a flux contribution
\(-\nabla\!\cdot\!\Psi(F_t)\) and a boundary contribution
\(M_{\partial\Omega}\odot\chi(F_t)\).
Here, \(\Omega\) is the spatial domain and \(\Psi\)
predicts a flux. Its divergence measures net outward
transport, so the negative divergence gives the
associated local state correction.
The mask \(M_{\partial\Omega}\) restricts the correction
predicted by \(\chi\) to the boundary region;
\(\odot\) denotes elementwise multiplication.

\subsection{Update}

The Update advances the current field by combining
the base increment with the closure correction:
\begin{equation}
\hat{x}_{t+1}
=
x_t+\Delta_{\mathrm{base}}(x_t)+\Delta_{\mathrm{clos}}.
\label{eq:model_update}
\end{equation}
This additive form follows the closure formulation
in the Introduction. The base increment describes
evolution inferred from the current field, and the
closure correction incorporates commutator evidence
and trajectory history.
The prediction becomes the Current State for the next
rollout step, together with the updated memory
\(h_{t+1}\).
Each correction therefore also changes the field from
which later commutator evidence is computed.

\subsection{Training Objective}
\label{sec:objective}

The training objective links each correction to both
current prediction accuracy and its effect on later steps.
All modules are optimized jointly through autoregressive
rollouts.
The objective combines prediction, commutator
regularization, and auxiliary regularization:
\begin{equation}
\mathcal{L}
=
\mathcal{L}_{\mathrm{pred}}
+\mathcal{L}_{\mathrm{comm}}
+\mathcal{L}_{\mathrm{reg}}.
\label{eq:training_objective}
\end{equation}
Each term includes its corresponding weights.
\(\mathcal{L}_{\mathrm{pred}}\) measures rollout
prediction error in physical space and the mismatch
between predicted and reference Fourier amplitudes.
The physical-space term trains field accuracy, while
the Fourier term penalizes differences in the strength
of spatial frequencies.
\(\mathcal{L}_{\mathrm{comm}}\) penalizes the squared,
unnormalized state and memory discrepancies of the
provisional paths across scales.
This soft penalty balances cross-scale agreement
against rollout accuracy, while the normalized
discrepancies serve as evidence for the final update.
\(\mathcal{L}_{\mathrm{reg}}\) regularizes memory
magnitude and temporal variation, closure-increment
magnitude, expert diversity, and routing entropy. Appendix~\ref{sec:theory} provides theoretical analysis of CoMNO. Appendix~\ref{app:implementation_cost} details implementation and computational cost.

\section{Experiments}
\subsection{Experimental Setup}

\paragraph{Data and protocols.}
We evaluate CoMNO on five benchmarks from APEBench~\cite{koehler2024apebench} and RealPDEBench~\cite{hu2026realpdebench}, covering diverse nonlinear dynamics, flow geometries, and fluid--structure coupling.
APEBench includes 2D Burgers, Kuramoto--Sivashinsky (KS), and Kolmogorov flow, using native pseudo-spectral trajectories generated at the evaluated resolution.
RealPDEBench includes Cylinder Flow and Fluid--Structure Interaction (FSI), with numerical and real measurement data.
Across both suites, models receive four input frames and are trained to predict 20 future steps.
APEBench evaluation uses autoregressive rollouts of up to 197 steps, with the 100-step figures displaying the corresponding trajectory prefixes.
RealPDEBench evaluation extends predictions to 100 steps by recursively chaining five 20-step forecast blocks.
Normalization statistics are estimated separately for each modality using only its training split.

We compare CoMNO with six representative neural surrogates spanning complementary modeling approaches:
spectral F-FNO~\cite{tran2023factorized},
factorized-attention FactFormer~\cite{li2023factformer},
memory-augmented MemNO~\cite{ICLR2025_89379d5f},
diffusion-based PDE-Refiner~\cite{lippe2023pderefiner},
and pretrained DPOT~\cite{hao2024dpot} and Poseidon~\cite{herde2024poseidon}.
All baselines use official implementations and share the same data splits, channel conventions, normalization procedures, and training and evaluation protocols. 

\paragraph{Metrics.}
Let $\hat{Y},Y\in\mathbb{R}^{T\times C\times H\times W}$
denote predicted and reference trajectories on the evaluation
grid, where $C$ counts the evaluated channels.
All metrics use inverse-normalized fields and are averaged
over test trajectories. Official RealPDEBench evaluation
excludes trailing zero-filled channels for unmeasured variables.

Rel.\ $\ell_2$ measures full-trajectory error as
$\|\hat{Y}-Y\|_2/\|Y\|_2$, with norms over time, channels,
and space. For stepwise evaluation, we define
\begin{equation}
e_t=
\frac{\|\hat{Y}_t-Y_t\|_2}{\|Y_t\|_2+\epsilon},
\label{eq:stepwise_nrmse}
\end{equation}
where $\epsilon=10^{-6}$ and norms jointly cover channels
and spatial locations.
Trajectory nRMSE averages $e_t$ over the forecast window,
while Final Step nRMSE reports $e_T$.
Mean Corr.\ averages the stepwise Pearson correlations
$\operatorname{cov}(\hat{Y}_t,Y_t)/
(\sigma_{\hat{Y}_t}\sigma_{Y_t}+10^{-8})$,
with covariance and population standard deviations computed
over the same flattened channel and spatial entries.

HF Error measures relative complex Fourier error within the
radial frequency band $\Omega_{\mathrm{high}}$ spanning
one-third to two-thirds of the Nyquist frequency:
\begin{equation}
\mathrm{HF\ Error}
=
\frac{1}{T}\sum_{t=1}^{T}
\frac{
\left(
\sum_{c=1}^{C}\sum_{k\in\Omega_{\mathrm{high}}}
w_k
\left|
\mathcal{F}(\hat{Y}_{t,c})_k-\mathcal{F}(Y_{t,c})_k
\right|^2
\right)^{1/2}
}{
\left(
\sum_{c=1}^{C}\sum_{k\in\Omega_{\mathrm{high}}}
w_k\left|\mathcal{F}(Y_{t,c})_k\right|^2
\right)^{1/2}+\epsilon
}.
\label{eq:hf}
\end{equation}
Here, $\mathcal{F}$ is the orthonormal spatial 2D real FFT;
the Nyquist frequency is $1/2$ cycles per grid point.
The Parseval weight $w_k$ equals one on the zero-frequency
and, when present, Nyquist columns of the reduced FFT axis,
and two elsewhere.

We further report four physical-consistency diagnostics
on the same evaluation grid.
PDE Step Residual (StepRes) for Burgers and Vorticity Relative
$\ell_2$ Error (VortRel) for Cylinder are defined as
\begin{equation}
\mathrm{StepRes}
=
\frac{1}{T-1}\sum_{t=1}^{T-1}
\frac{\|\hat{Y}_{t+1}-\Phi_{\mathrm{ref}}(\hat{Y}_t)\|_{\mathrm{rms}}}
{\max\{\|\Phi_{\mathrm{ref}}(\hat{Y}_t)\|_{\mathrm{rms}},
\epsilon_{\mathrm{phys}}\}},
\qquad
\mathrm{VortRel}
=
\frac{1}{T}\sum_{t=1}^{T}
\frac{\|\hat{\omega}_t-\omega_t\|_2}
{\|\omega_t\|_2+\epsilon}.
\label{eq:physical_field_metrics}
\end{equation}
Here, $\Phi_{\mathrm{ref}}$ is the APEBench reference time-stepper;
$\|\cdot\|_{\mathrm{rms}}$ denotes RMS over components and space,
with $\epsilon_{\mathrm{phys}}$ set to $10^{-12}$.
Cylinder vorticity $\omega_t$ is the spatial curl of the
velocity channels in $Y_t$, with $\hat{\omega}_t$ computed
analogously from $\hat{Y}_t$.
Norms cover all relevant components and spatial locations.
StepRes measures consistency with reference numerical evolution,
whereas VortRel measures error in the spatial vorticity field.

Energy Curve Relative Error for Burgers divides the temporal
RMS of the prediction-reference energy difference by the
reference energy curve's RMS, floored at $\epsilon_{\mathrm{phys}}$.
Energy averages squared field values over components and space.
Enstrophy Relative Error for Cylinder averages stepwise absolute
differences in spatial mean squared vorticity, divided by
corresponding reference values plus $\epsilon$.
These errors measure fidelity to reference energy and enstrophy
evolution, without assuming either quantity is conserved.

\begin{table*}[!htbp]
\centering
\scriptsize
\setlength{\tabcolsep}{3.8pt}
\renewcommand{\arraystretch}{0.92}
\caption{
Main comparison on long-horizon PDE rollouts for Burgers from APEBench~\cite{koehler2024apebench} and Cylinder Flow from RealPDEBench~\cite{hu2026realpdebench}.
}

\resizebox{\textwidth}{!}{%
\begin{tabular}{lccccc}
\toprule
Model
& Rel. $\ell_2$ $\downarrow$
& Trajectory nRMSE $\downarrow$
& Final Step nRMSE $\downarrow$
& HF Error $\downarrow$
& Mean Corr. $\uparrow$ \tabularnewline
\midrule

\multicolumn{6}{c}{
\emph{Burgers Equation from APEBench~\cite{koehler2024apebench}}
} \tabularnewline
\midrule

DPOT~\cite{hao2024dpot}
& $0.3959$
& $0.3959$
& $0.6571$
& $25.2052$
& $0.9051$ \tabularnewline

FactFormer~\cite{li2023factformer}
& $0.4442$
& $0.4442$
& $0.6559$
& $4.1411$
& $0.8814$ \tabularnewline

F-FNO~\cite{tran2023factorized}
& $0.0969$
& $0.0969$
& $0.1555$
& $0.5222$
& $0.9946$ \tabularnewline

MemNO~\cite{ICLR2025_89379d5f}
& $0.2139$
& $0.1548$
& $0.4255$
& $8.1969$
& $0.9728$ \tabularnewline

PDE-Refiner~\cite{lippe2023pderefiner}
& $0.4152$
& $0.0698$
& $0.8458$
& $13.8931$
& $0.8947$ \tabularnewline

Poseidon~\cite{herde2024poseidon}
& $0.8926$
& $0.6306$
& $0.4520$
& $1.4260$
& $0.8935$ \tabularnewline

\textbf{Ours}
& $\mathbf{0.0568}$
& $\mathbf{0.0568}$
& $\mathbf{0.1412}$
& $\mathbf{0.4688}$
& $\mathbf{0.9978}$ \tabularnewline

\midrule

\multicolumn{6}{c}{
\emph{Cylinder Flow from RealPDEBench~\cite{hu2026realpdebench}}
} \tabularnewline
\midrule

DPOT~\cite{hao2024dpot}
& $0.1389$
& $0.1861$
& $0.2066$
& $0.9526$
& $0.9795$ \tabularnewline

FactFormer~\cite{li2023factformer}
& $0.1242$
& $0.2046$
& $0.1730$
& $0.8092$
& $0.9564$ \tabularnewline

F-FNO~\cite{tran2023factorized}
& $0.1302$
& $0.1302$
& $0.1815$
& $0.8115$
& $0.9845$ \tabularnewline

MemNO~\cite{ICLR2025_89379d5f}
& $0.1282$
& $0.1282$
& $0.1807$
& $0.8546$
& $0.9854$ \tabularnewline

PDE-Refiner~\cite{lippe2023pderefiner}
& $0.1248$
& $0.1248$
& $0.1829$
& $0.8150$
& $0.9855$ \tabularnewline

Poseidon~\cite{herde2024poseidon}
& $0.1370$
& $0.1370$
& $0.2020$
& $0.9568$
& $0.9802$ \tabularnewline

\textbf{Ours}
& $\mathbf{0.1181}$
& $\mathbf{0.1181}$
& $\mathbf{0.1718}$
& $\mathbf{0.8080}$
& $\mathbf{0.9874}$ \tabularnewline

\bottomrule
\end{tabular}
}
\label{tab:main_comparison_combined}

\end{table*}

\begin{figure}[!t]
\centering

\begin{minipage}[t]{0.49\textwidth}
    \centering
    \includegraphics[
        width=\linewidth,
        keepaspectratio
    ]{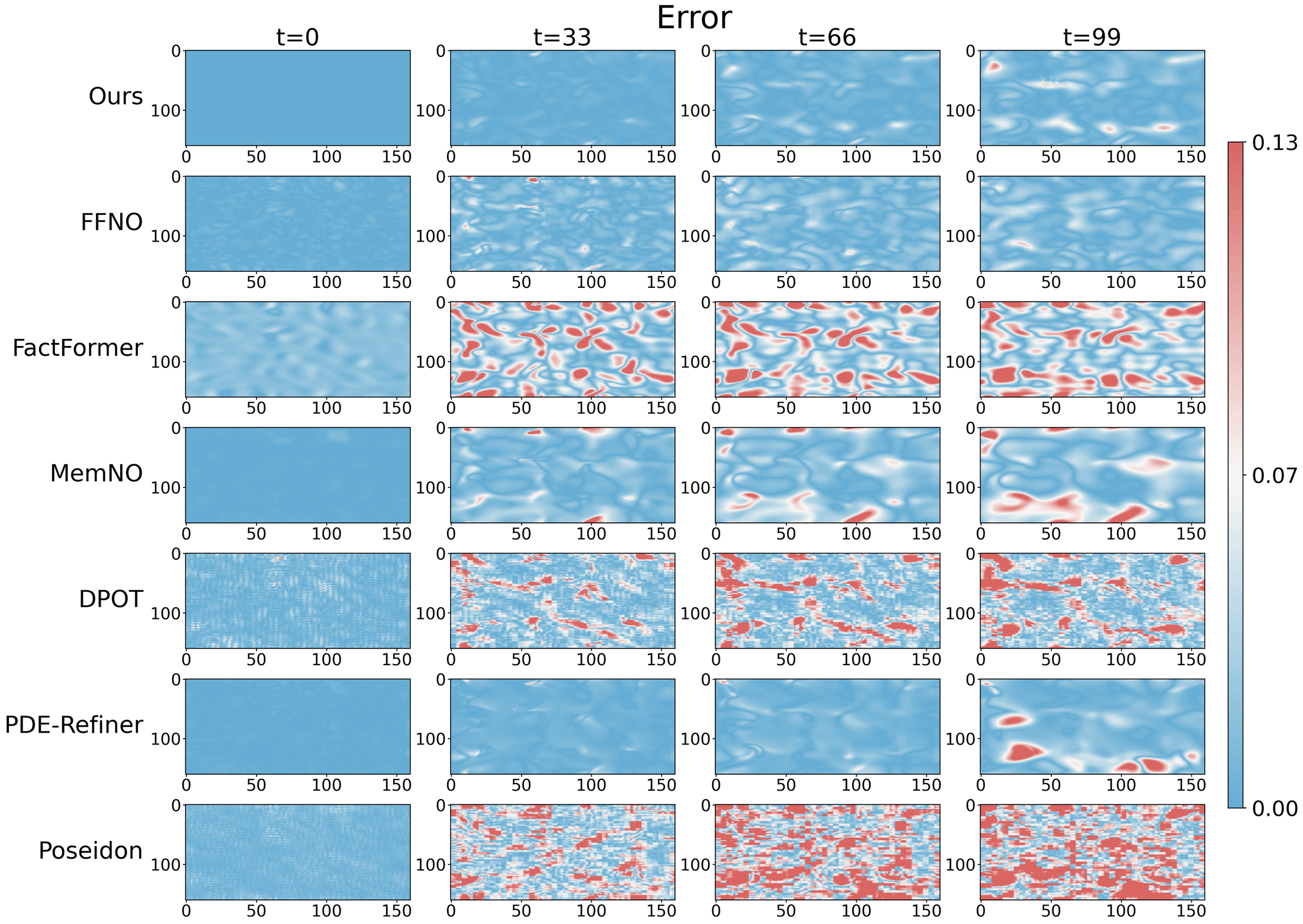}
    \captionof{figure}{
    Error evolution over long-horizon rollouts on APEBench Burgers~\cite{koehler2024apebench}.
    }
    \label{fig:burgers_error}
\end{minipage}
\hfill
\begin{minipage}[t]{0.49\textwidth}
    \centering
    \includegraphics[
        width=\linewidth,
        keepaspectratio
    ]{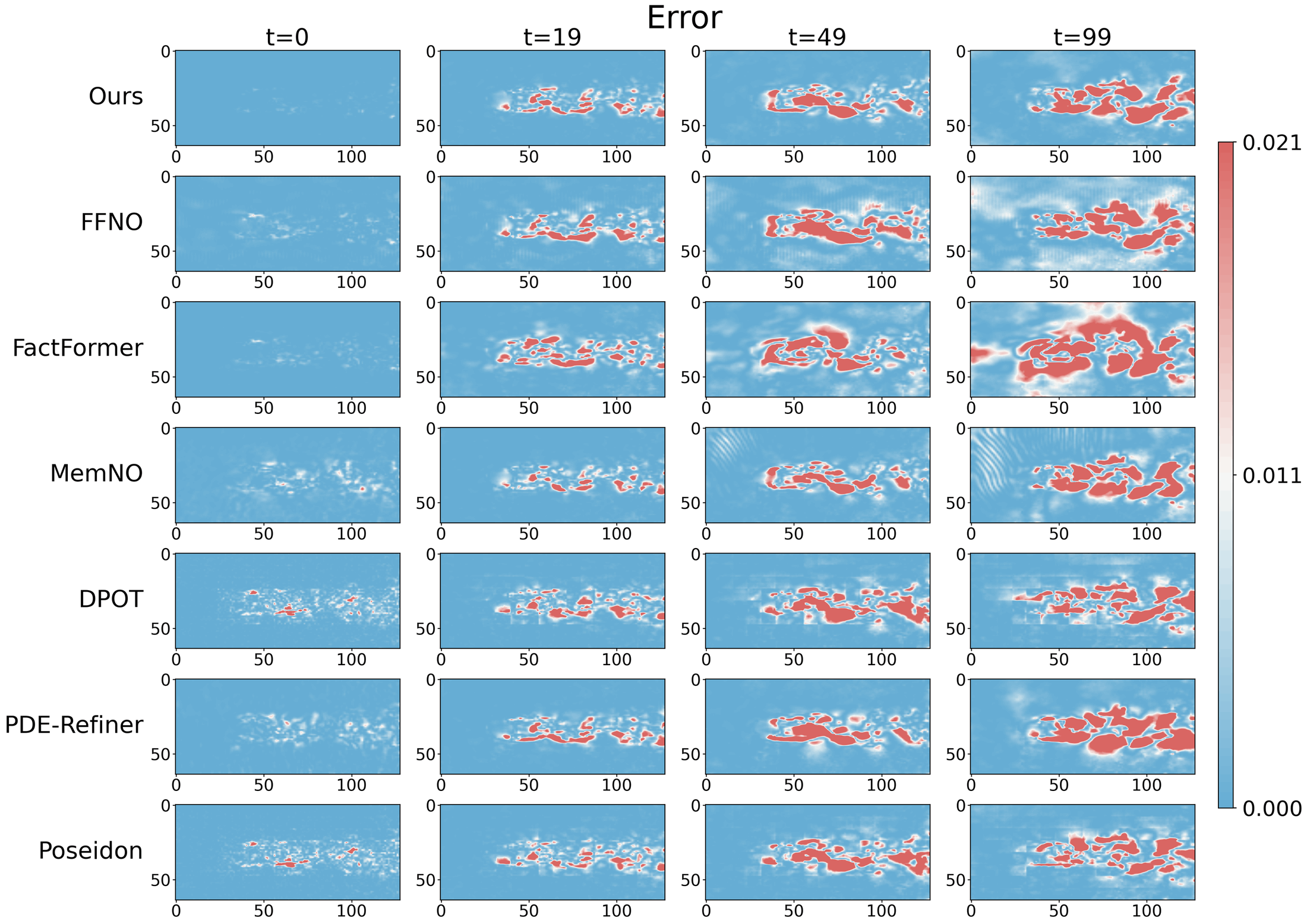}
    \captionof{figure}{
    Error evolution over long-horizon rollouts on RealPDEBench Cylinder Flow~\cite{hu2026realpdebench}.
    }
    \label{fig:cylinder_errorr}
    \vspace{-1.8em}
\end{minipage}
\end{figure}

\subsection{Main Comparison}
\label{sec:main_comparison}

Table~\ref{tab:main_comparison_combined} compares CoMNO with representative
neural operators on APEBench Burgers~\cite{koehler2024apebench} and
RealPDEBench Cylinder Flow~\cite{hu2026realpdebench}. CoMNO achieves the
best reported mean on every metric across both benchmarks. Its advantages
span trajectory-level and terminal accuracy, accompanied by improvements
in high-frequency fidelity and spatial correlation. These complementary
results demonstrate that CoMNO improves long-horizon prediction accuracy
while more faithfully capturing the spectral content and spatial
structure of the predicted fields.

Figures~\ref{fig:burgers_error} and~\ref{fig:cylinder_errorr} further
illustrate the evolution of spatial errors during autoregressive
rollouts. In particular, CoMNO exhibits predominantly low error at the
displayed Burgers prediction times, whereas several baselines develop
pronounced deviations in later snapshots, consistent with the
quantitative comparison. Additional quantitative results and
visualizations are provided in Appendix~\ref{sec:main_supp}.

\subsection{Functional Role of the Commutator Signal}
\label{subsec:commutator_diagnostics}

We intervene on frozen checkpoints via
$\bar e_t \mapsto \lambda \bar e_t$, comparing signal removal
($\lambda=0$) with the unmodified predictor ($\lambda=1$).
All learned parameters remain fixed, and the signal is recomputed
along each resulting trajectory.
Let $\varepsilon_{i,m}^{(\lambda)}$ denote the relative $\ell_2$
error of $\hat x_t^{(\lambda)}$ against $x_t$ under metric $m$.
We report the mean paired gain
\begin{equation}
\overline G_m
=
\frac{1}{N}\sum_{i=1}^{N}
\left(
\varepsilon_{i,m}^{(0)}-\varepsilon_{i,m}^{(1)}
\right),
\end{equation}
where $i$ indexes independent trajectories or physical trajectory
groups. The metrics evaluate the first prediction, average error
over the full rollout and its final quarter, and full-rollout
error restricted to Fourier modes above the median radial
wavenumber (HF).

Table~\ref{tab:commutator_gain} reveals a consistent
pattern across systems: restoring the signal yields small
one-step gains, larger full-rollout gains, and late-horizon
gains that consistently exceed the full-rollout average.
The signal's predictive contribution is therefore more fully
expressed during closed-loop evolution, where its influence
propagates through autoregressive feedback.
Positive HF gains across all systems further show that the
benefit extends to fine-scale spatial structure.
These interventions at fixed parameters establish the
commutator's functional utility within the trained predictor
and support its role as an informative input for online
closure correction. Appendix~\ref{app:module3_commutator_diagnostics} shows that the commutator improves terminal prediction, evolves with the trajectory, and benefits from calibrated signal strength.

\begin{figure*}[t]
\centering

\begin{minipage}[t]{0.49\textwidth}
    \vspace{0pt}
    \centering

    \captionof{table}{
    Mean paired error reductions from restoring the commutator signal; positive values indicate improvement.
    }
    \label{tab:commutator_gain}
    \footnotesize
    \setlength{\tabcolsep}{4pt}
    \renewcommand{\arraystretch}{1.12}

    \resizebox{\linewidth}{!}{%
    \begin{tabular}{lcccc}
    \toprule
    Dataset
    & \shortstack{One-step\\gain $\uparrow$}
    & \shortstack{Full-rollout\\gain $\uparrow$}
    & \shortstack{Late-horizon\\gain $\uparrow$}
    & \shortstack{HF\\gain $\uparrow$} \\
    \midrule
    Burgers    & 0.0001 & 0.0045 & 0.0098 & 0.0105 \\
    KS         & 0.0001 & 0.0101 & 0.0118 & 0.0233 \\
    Kolmogorov & 0.0002 & 0.0178 & 0.0291 & 0.0345 \\
    Cylinder   & 0.0003 & 0.0487 & 0.0679 & 0.0503 \\
    FSI        & 0.0001 & 0.0084 & 0.0137 & 0.0195 \\
    \bottomrule
    \end{tabular}%
    }
\end{minipage}
\hfill
\begin{minipage}[t]{0.49\textwidth}
    \vspace{0pt}
    \centering

    \captionof{table}{
    Relative error change (\%) against the best baseline per setting and metric; negative values indicate improvement.
    }
    \label{tab:coarse_gain}

    \footnotesize
    \setlength{\tabcolsep}{4pt}
    \renewcommand{\arraystretch}{1.12}

    \resizebox{\linewidth}{!}{%
    \begin{tabular}{lcccc}
    \toprule
    Dataset & Factor & Mean nRMSE & Final nRMSE & Spectral Error \\
    \midrule
    Burgers    & $4\times$ & $-53.3$  & $-57.9$  & $-24.3$ \\
    Burgers    & $8\times$ & $-4.2$ & $-44.1$  & $+22.7$  \\
    \midrule
    KS         & $4\times$ & $-0.3$  & $-0.4$  & $-2.5$ \\
    KS         & $8\times$ & $-0.6$   & $-17.0$  & $-7.5$  \\
    \midrule
    Kolmogorov & $4\times$ & $-21.2$   & $-0.6$ & $-14.4$  \\
    Kolmogorov & $8\times$ & $-0.3$   & $-4.0$  & $-0.9$  \\
    \bottomrule
    \end{tabular}%
    }
\end{minipage}

\end{figure*}

\subsection{Performance under Spatial Coarsening}
\label{sec:coarsening}

We evaluate CoMNO under $4\times$ and $8\times$ spatial
coarsening to test closure correction with reduced spatial
information. Table~\ref{tab:coarse_gain} shows improvements
over the best baseline in 17 of 18 comparisons, spanning
trajectory accuracy, terminal accuracy, and spectral
fidelity. On Burgers, terminal accuracy improves markedly
at both coarsening levels, while FactFormer leads spectral
accuracy at $8\times$. Gains vary
across systems and metrics. Together with the commutator
interventions, the results support CoMNO's central design:
extracting closure evidence from observable cross-scale
discrepancies and retaining its temporal context to guide
predictions when fine-scale structures are unresolved. Appendix~\ref{app:coarse_observation} reports the full coarsening results under $4\times$ and $8\times$ spatial coarsening.

\subsection{Rollout Stability}
\label{sec:rollout}

Figure~\ref{fig:rollout} evaluates autoregressive stability on Burgers
and Cylinder flow using relative \(\ell_2\) error across rollout
steps and final nRMSE at the last evaluated step.
CoMNO achieves the lowest relative \(\ell_2\) error at every
displayed horizon on both systems, maintaining its advantage
as predictions are recursively reused as inputs.
On Burgers, it remains clearly separated from most baselines
at later steps.
On Cylinder flow, all methods exhibit increasing error,
with CoMNO consistently tracing the lower envelope.
The insets corroborate this advantage through the lowest
final nRMSE at step \(197\) for Burgers and step \(100\)
for Cylinder.
Together, these results demonstrate sustained predictive
accuracy under extended autoregressive rollout.
This persistence is consistent with CoMNO's use of commutator
evidence and Closure Memory to inform successive closure
corrections as the trajectory evolves.
Full statistics and extended-horizon results are reported
in Appendix~\ref{sec:rollout_supp}.

\subsection{Physical Consistency}
\label{sec:physical_consistency}

Figure~\ref{fig:physics} complements the rollout evaluation
with four physical diagnostics.
CoMNO achieves the lowest reported values on both axes
in each system.
On Burgers, it combines the lowest StepRes with the most
accurate energy evolution. On Cylinder flow, it jointly
reduces vorticity and enstrophy errors, improving agreement
with the reference in both the spatial flow structure
and its aggregate rotational intensity.
The strongest competing baseline differs between the two
systems, yet CoMNO retains its advantage on both diagnostic
pairs.
These gains show that CoMNO's sustained rollout accuracy
is accompanied by more faithful energy and vorticity
evolution, with improvements extending from spatial
fields to global physical observables.
Together, the results support improved physical fidelity
alongside lower prediction error.
Full statistics and system-specific analyses are provided
in Appendix~\ref{sec:physics_supp}.

\subsection{Ablation Study}
\label{sec:ablation}

\begin{figure*}[t]
    \centering

    \begin{minipage}[t]{0.49\textwidth}
        \centering
        \includegraphics[
            width=\linewidth,
            keepaspectratio
        ]{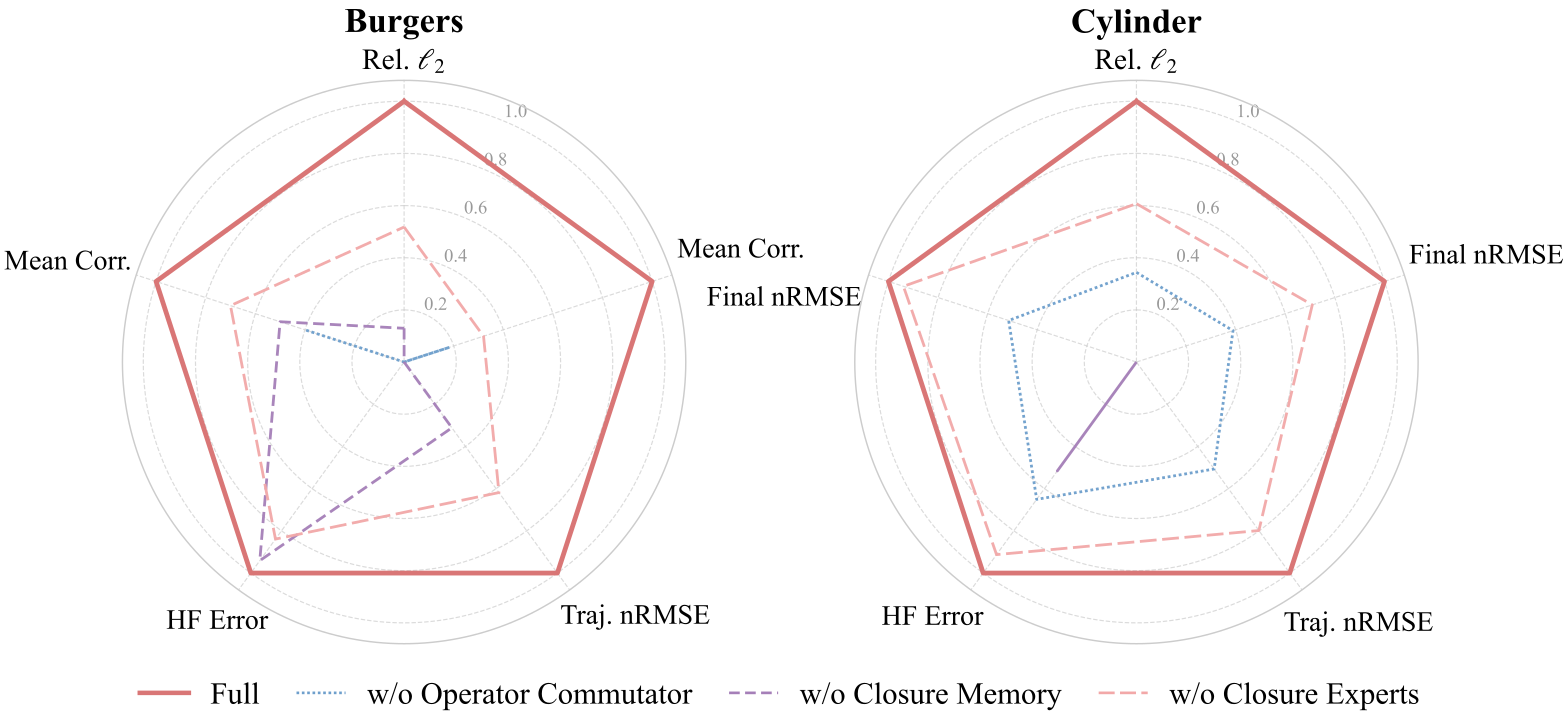}
        \captionof{figure}{
        Ablation on Burgers and Cylinder.
        }
        \label{fig:ablation}
    \end{minipage}
    \hfill
    \begin{minipage}[t]{0.49\textwidth}
        \centering
        \includegraphics[
            width=\linewidth,
            keepaspectratio
        ]{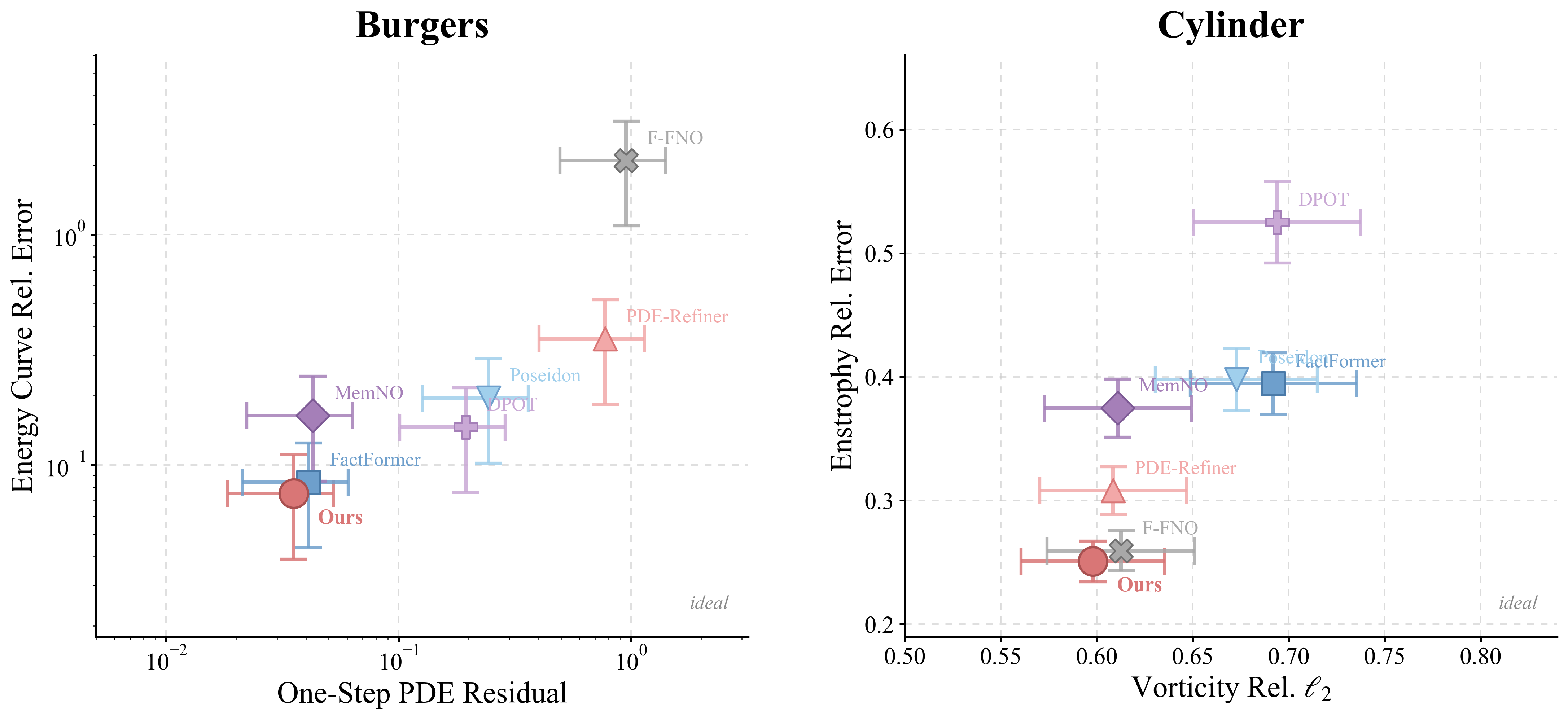}
        \captionof{figure}{
        Physical consistency on Burgers and Cylinder.
        }
        \label{fig:physics}
    \end{minipage}

\end{figure*}

Figure~\ref{fig:ablation} evaluates CoMNO by individually
removing Operator Commutator, Closure Memory, and Closure
Experts. The full model performs best across all five
metrics on both Burgers and Cylinder flow.
Each removal degrades performance across multiple aspects
of prediction, affecting state accuracy, rollout error,
spectral fidelity, and spatial correlation.
The losses in both trajectory and final-step accuracy
show that the contribution of each module extends
through the rollout to its endpoint.
Although the effects vary across metrics and systems,
retaining all three modules consistently yields the
strongest overall performance.
These results support the combined use of commutator
evidence, temporal memory, and adaptive closure correction:
each module contributes to the full model's performance,
with benefits spanning temporal prediction and spatial
fidelity.
Complete quantitative results and further analyses
are provided in Appendix~\ref{sec:ablation_supp}.

\begin{figure*}[t]
    \centering
    \includegraphics[
        width=\textwidth,
        keepaspectratio
    ]{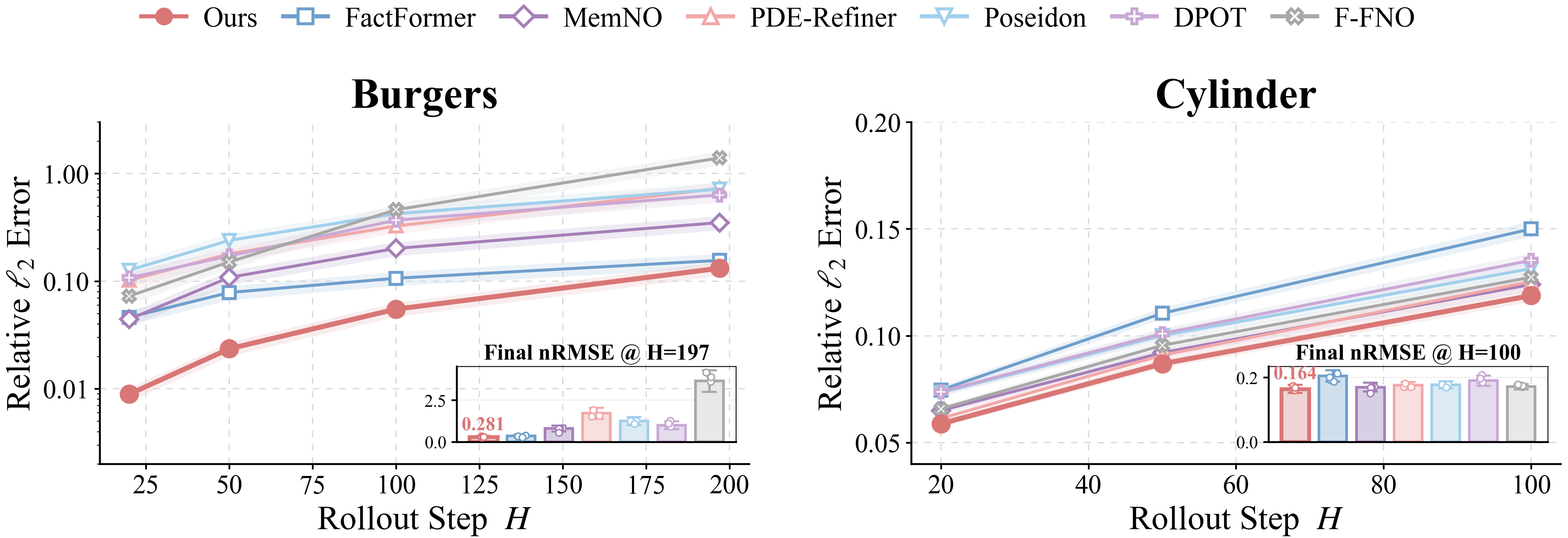}
    \caption{
    Rollout error across steps on Burgers and Cylinder.
    }
    \label{fig:rollout}
\end{figure*}

\section{Conclusion}

We presented CoMNO, which learns history-dependent closure
by combining cross-scale evidence with trajectory memory.
The Operator Commutator measures discrepancies between
evolution and downsampling orders; Closure Memory retains
their history; and Closure Experts generate spatially
adaptive corrections.
Across five PDE systems, CoMNO improves predictive accuracy,
rollout stability, spectral fidelity, and physical consistency.
Gains extend beyond the training horizon and persist in most
comparisons under spatial coarsening.
Frozen-checkpoint interventions reveal stronger commutator
benefits at later horizons, while ablations support the
core components.
Together, these results demonstrate the value of integrating
current cross-scale discrepancies with their temporal history
for accurate long-horizon PDE prediction.

\section*{AI Use Statement}

AI tools assisted with research ideation and experimental design, drafting portions of the manuscript, and refining the writing. The authors reviewed and verified all AI-assisted content, including technical claims and experimental procedures, and take full responsibility for the final manuscript and reported results.

\bibliography{iclr2027_conference}
\bibliographystyle{iclr2027_conference}

\appendix

\section{Theory of Commutator-Guided Closure}
\label{sec:theory}

We analyze how the Operator Commutator supplies a residual proxy, when
the memory-conditioned correction improves prediction, and how its
effect propagates through autoregressive rollout. Model parameters
are fixed. We retain the notation of the main text and use
$\|\cdot\|_2$ for the Euclidean norm over field components and spatial
locations.

\subsection{Calibration of the Online Commutator}
\label{subsec:theory_commutator}

The two provisional paths differ in the order of prediction and
downsampling. Path A predicts on the observation grid before
downsampling, whereas Path B first downsamples the state and memory.
Both use $\Phi^0$ with zero current evidence. Their discrepancy
provides an estimate of the correction needed by Path B.

To make this interpretation precise, let $r_t^{(s)}$ denote the
correction that makes the Path B prediction equal the restricted
reference state $D_sx_{t+1}$. The commutator and this correction
subtract the same Path B prediction. Their difference is therefore
exactly the observation-grid provisional prediction error after
restriction. For linear $D_s$, this gives
\begin{equation}
\bigl\|\tilde c_t^{(s)}-r_t^{(s)}\bigr\|_2
\leq
\|D_s\|_{2\to2}
\bigl\|\Phi^0(x_t;h_t)-x_{t+1}\bigr\|_2,
\label{eq:theory_commutator_bound}
\end{equation}
where $\|D_s\|_{2\to2}$ is the induced operator norm.
The bound connects the accuracy of the residual estimate directly
to the provisional forecast error. In particular, an exact forecast
on the observation grid makes the commutator equal the correction
required on the restricted grid.

CoMNO normalizes, lifts, concatenates, and averages the state and
memory discrepancies across scales to form $\bar e_t$. Starting
again from the same pair $(x_t,h_t)$, the final pass uses this
processed evidence together with local features and retained
history. Closure Experts thereby learn how cross-resolution
disagreement informs the correction on the observation grid.

\subsection{Memory-Conditioned Closure and the Value of Evidence}
\label{subsec:theory_closure}

All fields in this subsection are square-integrable under the
reference-trajectory distribution. The available history
$\mathcal H_t$ includes $x_t$, and $h_t$ is obtained causally from this
history using the model's initialization and memory updates. Following
the main-text formulation, the target supplied to the learned closure is
\begin{equation}
R_t:=x_{t+1}-x_t-\Delta_{\mathrm{base}}(x_t).
\label{eq:theory_closure_target}
\end{equation}
Under the closure decomposition in the main text,
$\Delta_{\mathrm{base}}$ implements the resolved increment, so $R_t$
is the history-dependent residual that Closure Experts approximate.
Its conditional means with and without memory are
\begin{equation}
m_t^h:=\mathbb E[R_t\mid x_t,h_t],
\qquad
m_t^x:=\mathbb E[R_t\mid x_t].
\label{eq:theory_closure_means}
\end{equation}

\begin{samepage}
\begin{proposition}[Retained information and approximation error]
\label{prop:theory_closure_gain}
For the final CoMNO prediction from the reference pair $(x_t,h_t)$,
\begin{equation}
\begin{aligned}
&\mathbb E\|R_t-m_t^x\|_2^2
 -\mathbb E\|x_{t+1}-\hat x_{t+1}\|_2^2\\
&\qquad=\mathbb E\|m_t^h-m_t^x\|_2^2
 -\mathbb E\|m_t^h-\Delta_{\mathrm{clos}}\|_2^2.
\end{aligned}
\label{eq:theory_closure_gain}
\end{equation}
\end{proposition}

\begin{proof}
The model update gives
$x_{t+1}-\hat x_{t+1}=R_t-\Delta_{\mathrm{clos}}$.
At fixed parameters, $\bar e_t$, $F_t$, and
$\Delta_{\mathrm{clos}}$ are functions of $(x_t,h_t)$.
Hence $R_t-m_t^h$ is orthogonal in mean square to both
$m_t^h-m_t^x$ and $m_t^h-\Delta_{\mathrm{clos}}$.
Decompose $R_t-m_t^x$ and $R_t-\Delta_{\mathrm{clos}}$ through
$m_t^h$, expand their squared norms, and subtract. The common residual
variance cancels.
\end{proof}
\end{samepage}

The left side compares the optimal state-only prediction risk with
CoMNO's risk. The right side separates the predictive variation resolved
by memory from the learned closure's approximation error. The quantity
$m_t^h-m_t^x$ therefore measures the part of the history-dependent
residual that becomes predictable once retained context is available.
CoMNO improves on the optimal state-only forecast precisely when its
closure approximation error is smaller than this retained-information
benefit.

The Router's softmax weights make $\Delta_{\mathrm{clos}}$ a convex
combination of $E_k(F_t)$ at each location. Its squared error relative
to $m_t^h$ is therefore at most the weighted average of expert errors.
Spatial summation and expectation bound the approximation term in
Eq.~\eqref{eq:theory_closure_gain}.

\paragraph{Gain from the actual evidence-conditioned correction.}
At the same reference pair $(x_t,h_t)$, let
$\Delta_{\mathrm{clos}}^{(\lambda)}$ be the correction produced with
current evidence $\lambda\bar e_t$, for $\lambda\in\{0,1\}$.
Features, routing weights, and expert outputs are evaluated with that
evidence, while the starting state and memory remain fixed. Write
$\delta_{\mathrm{clos}}:=
\Delta_{\mathrm{clos}}^{(1)}-\Delta_{\mathrm{clos}}^{(0)}$.
Both passes approximate the same conditional mean $m_t^h$, and their
mean-square prediction gain is exactly
\begin{equation}
\begin{aligned}
&\mathbb E\|R_t-\Delta_{\mathrm{clos}}^{(0)}\|_2^2
 -\mathbb E\|R_t-\Delta_{\mathrm{clos}}^{(1)}\|_2^2\\
&\qquad=
2\mathbb E\langle m_t^h-\Delta_{\mathrm{clos}}^{(0)},
                    \delta_{\mathrm{clos}}\rangle
-\mathbb E\|\delta_{\mathrm{clos}}\|_2^2.
\end{aligned}
\label{eq:theory_local_evidence_gain}
\end{equation}
Expanding the squared norms and conditioning on $(x_t,h_t)$ replaces
$R_t$ by $m_t^h$ in the cross term. Thus, evidence improves the actual
one-step mean-square prediction
precisely when twice the expected alignment with the remaining mean
residual exceeds the expected squared correction norm. Closure Memory
supplies retained historical information; the current evidence guides
the learned approximation of the resulting mean closure.

\subsection{Closed-Loop Propagation and Prediction Gains}
\label{subsec:theory_rollout}

We now apply the frozen-checkpoint intervention
$\bar e_t\mapsto\lambda\bar e_t$ throughout a rollout, with a common
initial field and memory. As in the main text, each rollout updates
both state and memory and recomputes evidence from its own inputs
according to the fixed evaluation schedule. In this subsection, state
fields are expressed after the inverse normalization used for evaluation.

\paragraph{Propagation through state and memory.}
Let $\hat x_T^{[j]}$ be the terminal prediction when evidence is removed
for the first $j$ updates and enabled thereafter. All these hybrid
rollouts start from the same initialization. Their endpoints are
$\hat x_T^{[0]}=\hat x_T^{(1)}$ and
$\hat x_T^{[T]}=\hat x_T^{(0)}$, so telescoping gives
\begin{equation}
\delta_T
:=\hat x_T^{(1)}-\hat x_T^{(0)}
=\sum_{j=0}^{T-1}
  \bigl(\hat x_T^{[j]}-\hat x_T^{[j+1]}\bigr).
\label{eq:theory_rollout_response}
\end{equation}
Adjacent hybrids share the same state--memory input at update $j$ and
differ in whether evidence is supplied at that update. Each terminal
contrast includes the subsequent nonlinear response of the state,
memory, and recomputed evidence. A change to the immediate memory output
can therefore affect later states even when the immediate state
predictions coincide.

\par
\noindent\begin{minipage}{\linewidth}
\centering
\footnotesize
\setlength{\tabcolsep}{3.2pt}
\renewcommand{\arraystretch}{1.13}
\captionof{table}{Prediction accuracy on Burgers. Entries report means across runs.}
\label{tab:burgers_additional_metrics}
\sbox0{%
\begin{tabular}{lccccc}
\toprule
Model & Rel. $\ell_2$ $\downarrow$ & Traj. nRMSE $\downarrow$ & Final nRMSE $\downarrow$ & HF Error $\downarrow$ & Mean Corr. $\uparrow$ \\
\midrule
DPOT & $0.3959$ & $0.3959$ & $0.6571$ & $25.2052$ & $0.9051$ \\
FactFormer & $0.4442$ & $0.4442$ & $0.6559$ & $4.1411$ & $0.8814$ \\
F-FNO & $0.0969$ & $0.0969$ & $0.1555$ & $0.5222$ & $0.9946$ \\
MemNO & $0.2139$ & $0.1548$ & $0.4255$ & $8.1969$ & $0.9728$ \\
PDE-Refiner & $0.4152$ & $0.0698$ & $0.8458$ & $13.8931$ & $0.8947$ \\
Poseidon & $0.8926$ & $0.6306$ & $0.4520$ & $1.4260$ & $0.8935$ \\
\midrule
\textbf{CoMNO} & $\mathbf{0.0568}$ & $\mathbf{0.0568}$ & $\mathbf{0.1412}$ & $\mathbf{0.4688}$ & $\mathbf{0.9978}$ \\
\bottomrule
\end{tabular}
}
\ifdim\wd0>\linewidth
  \resizebox{\linewidth}{!}{\usebox0}%
\else
  \usebox0
\fi
\end{minipage}
\par
\noindent\begin{minipage}{\linewidth}
\centering
\footnotesize
\setlength{\tabcolsep}{3.2pt}
\renewcommand{\arraystretch}{1.13}
\captionof{table}{Prediction accuracy on Cylinder. Entries report means across runs.}
\label{tab:cylinder_additional_metrics}
\sbox0{%
\begin{tabular}{lccccc}
\toprule
Model & Rel. $\ell_2$ $\downarrow$ & Traj. nRMSE $\downarrow$ & Final nRMSE $\downarrow$ & HF Error $\downarrow$ & Mean Corr. $\uparrow$ \\
\midrule
DPOT & $0.1389$ & $0.1861$ & $0.2066$ & $0.9526$ & $0.9795$ \\
FactFormer & $0.1242$ & $0.2046$ & $0.1730$ & $0.8092$ & $0.9564$ \\
F-FNO & $0.1302$ & $0.1302$ & $0.1815$ & $0.8115$ & $0.9845$ \\
MemNO & $0.1282$ & $0.1282$ & $0.1807$ & $0.8546$ & $0.9854$ \\
PDE-Refiner & $0.1248$ & $0.1248$ & $0.1829$ & $0.8150$ & $0.9855$ \\
Poseidon & $0.1370$ & $0.1370$ & $0.2020$ & $0.9568$ & $0.9802$ \\
\midrule
\textbf{CoMNO} & $\mathbf{0.1181}$ & $\mathbf{0.1181}$ & $\mathbf{0.1718}$ & $\mathbf{0.8080}$ & $\mathbf{0.9874}$ \\
\bottomrule
\end{tabular}
}
\ifdim\wd0>\linewidth
  \resizebox{\linewidth}{!}{\usebox0}%
\else
  \usebox0
\fi
\end{minipage}
\par
\noindent\begin{minipage}{\linewidth}
\centering
\footnotesize
\setlength{\tabcolsep}{3.2pt}
\renewcommand{\arraystretch}{1.13}
\captionof{table}{Prediction accuracy on KS using a single forecast of the future sequence. Entries report means across runs.}
\label{tab:main_comparison_ks}
\sbox0{%
\begin{tabular}{lccccc}
\toprule
Model & Rel. $\ell_2$ $\downarrow$ & Traj. nRMSE $\downarrow$ & Final nRMSE $\downarrow$ & HF Error $\downarrow$ & Mean Corr. $\uparrow$ \\
\midrule
DPOT & $0.8902$ & $0.8902$ & $1.2258$ & $5.2418$ & $0.5141$ \\
FactFormer & $0.8176$ & $0.8176$ & $1.0127$ & $1.0503$ & $0.5062$ \\
F-FNO & $0.8435$ & $0.8435$ & $1.2171$ & $1.8693$ & $0.5537$ \\
MemNO & $0.3749$ & $0.1067$ & $0.8095$ & $1.3892$ & $0.9024$ \\
PDE-Refiner & $0.7620$ & $0.1140$ & $1.2125$ & $1.7985$ & $0.6464$ \\
Poseidon & $1.5387$ & $0.8356$ & $1.7151$ & $13.5822$ & $0.5183$ \\
\midrule
\textbf{CoMNO} & $\mathbf{0.1204}$ & $\mathbf{0.0903}$ & $\mathbf{0.2836}$ & $\mathbf{0.2226}$ & $\mathbf{0.9925}$ \\
\bottomrule
\end{tabular}
}
\ifdim\wd0>\linewidth
  \resizebox{\linewidth}{!}{\usebox0}%
\else
  \usebox0
\fi
\end{minipage}
\par

\paragraph{Connection to the reported error reductions.}
Define the terminal error field
$d_T^{(\lambda)}:=x_T-\hat x_T^{(\lambda)}$.
The intervention gives $d_T^{(1)}=d_T^{(0)}-\delta_T$.
Let $e_T^{(\lambda)}$ be the scalar stepwise relative error from the
main text, with the same $\epsilon=10^{-6}$. If at least one error
field is nonzero, expansion and the difference-of-squares identity give
\begin{equation}
e_T^{(0)}-e_T^{(1)}
=
\frac{
2\langle d_T^{(0)},\delta_T\rangle-\|\delta_T\|_2^2
}{
(\|x_T\|_2+\epsilon)
(\|d_T^{(0)}\|_2+\|d_T^{(1)}\|_2)
}.
\label{eq:theory_relative_gain}
\end{equation}
When both error fields vanish, the gain is zero. Otherwise, the
prediction improves precisely when twice the alignment of the total
response with the original error exceeds the squared response magnitude.
Longer-horizon gains depend on how the recurrent updates change both
this response and its alignment with the reference error.

Averaging these normalized differences over the prescribed times and
trajectory groups gives the corresponding paired gains $\overline G_m$.
For full-trajectory norms, stack the evaluated fields and use the
metric's normalization. For the intervention HF gain, apply the
above-median radial-frequency restriction defined in
Section~\ref{subsec:commutator_diagnostics}; the main HF Error retains
the one-third-to-two-thirds Nyquist band defined in the evaluation
metrics. The two spectral quantities are therefore distinct diagnostics.
The measured gains include the recurrent effect of evidence carried
forward by Closure Memory.

\par
\noindent\begin{minipage}{\linewidth}
\centering
\footnotesize
\setlength{\tabcolsep}{3.2pt}
\renewcommand{\arraystretch}{1.13}
\captionof{table}{Prediction accuracy on Kolmogorov flow. Entries report means across runs.}
\label{tab:main_comparison_kolm}
\sbox0{%
\begin{tabular}{lccccc}
\toprule
Model & Rel. $\ell_2$ $\downarrow$ & Traj. nRMSE $\downarrow$ & Final nRMSE $\downarrow$ & HF Error $\downarrow$ & Mean Corr. $\uparrow$ \\
\midrule
DPOT & $0.7963$ & $0.7910$ & $0.9136$ & $12.0998$ & $0.6113$ \\
FactFormer & $0.8143$ & $0.8143$ & $1.0123$ & $1.9624$ & $0.5773$ \\
F-FNO & $0.5784$ & $0.5784$ & $1.0111$ & $1.4050$ & $0.7794$ \\
MemNO & $0.8183$ & $0.5365$ & $1.0724$ & $5.6845$ & $0.5766$ \\
PDE-Refiner & $0.9183$ & $0.8337$ & $1.2231$ & $13.9558$ & $0.3836$ \\
Poseidon & $1.2515$ & $0.8961$ & $1.6569$ & $4.2207$ & $0.3730$ \\
\midrule
\textbf{CoMNO} & $\mathbf{0.4585}$ & $\mathbf{0.4585}$ & $\mathbf{0.8851}$ & $\mathbf{1.2676}$ & $\mathbf{0.8518}$ \\
\bottomrule
\end{tabular}
}
\ifdim\wd0>\linewidth
  \resizebox{\linewidth}{!}{\usebox0}%
\else
  \usebox0
\fi
\end{minipage}
\par
\noindent\begin{minipage}{\linewidth}
\centering
\footnotesize
\setlength{\tabcolsep}{3.2pt}
\renewcommand{\arraystretch}{1.13}
\captionof{table}{Prediction accuracy on FSI. Entries report means across runs.}
\label{tab:main_comparison_fsi}
\sbox0{%
\begin{tabular}{lccccc}
\toprule
Model & Rel. $\ell_2$ $\downarrow$ & Traj. nRMSE $\downarrow$ & Final nRMSE $\downarrow$ & HF Error $\downarrow$ & Mean Corr. $\uparrow$ \\
\midrule
DPOT & $0.1964$ & $0.2228$ & $0.4167$ & $0.9353$ & $0.9634$ \\
FactFormer & $0.1547$ & $0.1641$ & $0.2508$ & $\mathbf{0.6777}$ & $0.9822$ \\
F-FNO & $0.1793$ & $0.1793$ & $0.2991$ & $0.7158$ & $0.9757$ \\
MemNO & $0.1522$ & $0.1522$ & $0.2376$ & $\mathbf{0.6777}$ & $0.9831$ \\
PDE-Refiner & $0.1698$ & $0.1698$ & $0.2755$ & $0.7236$ & $0.9790$ \\
Poseidon & $0.2028$ & $0.2315$ & $0.4254$ & $0.9281$ & $0.9546$ \\
\midrule
\textbf{CoMNO} & $\mathbf{0.1487}$ & $\mathbf{0.1487}$ & $\mathbf{0.2305}$ & $0.6791$ & $\mathbf{0.9836}$ \\
\bottomrule
\end{tabular}
}
\ifdim\wd0>\linewidth
  \resizebox{\linewidth}{!}{\usebox0}%
\else
  \usebox0
\fi
\end{minipage}
\par
\noindent\begin{minipage}{\linewidth}
\begin{minipage}[t]{.485\linewidth}
    \centering
    \includegraphics[width=\linewidth,height=.215\textheight,keepaspectratio]{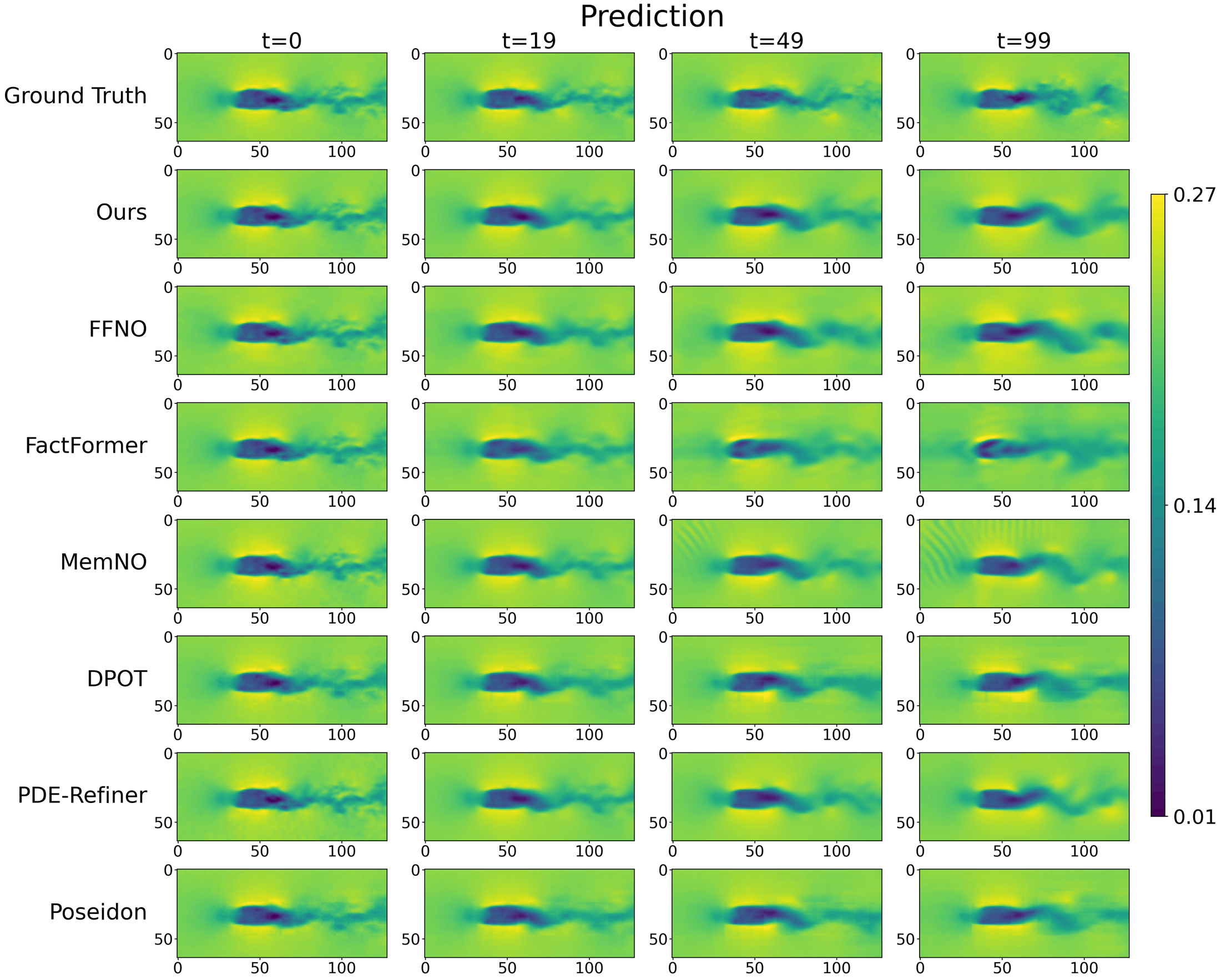}
    \captionof{figure}{Cylinder $u$ predictions}
    \label{fig:cylinder_prediction}
  \end{minipage}\hfill
  \begin{minipage}[t]{.485\linewidth}
    \centering
    \includegraphics[width=\linewidth,height=.215\textheight,keepaspectratio]{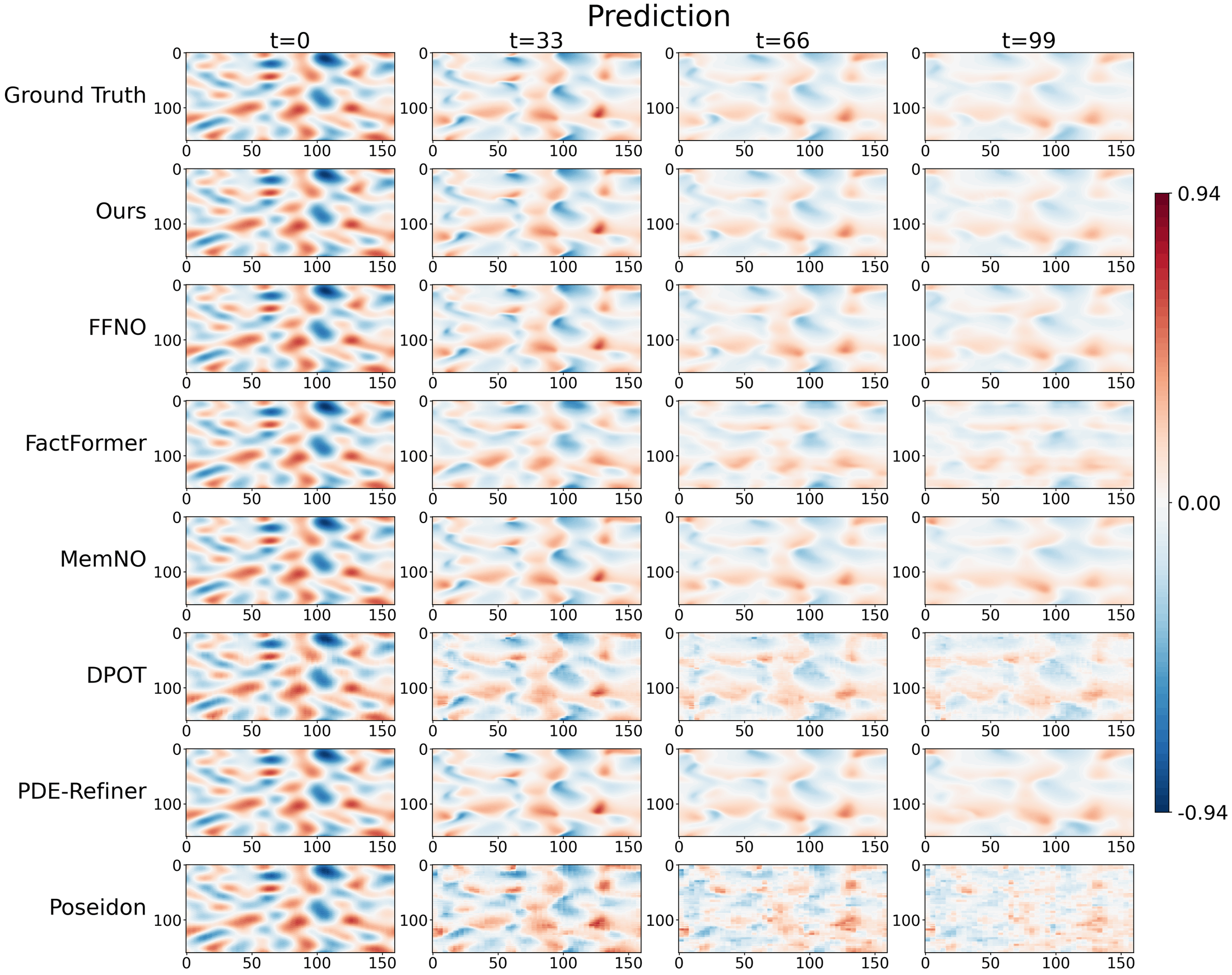}
    \captionof{figure}{Burgers predictions}
    \label{fig:burgers_prediction}
  \end{minipage}\par
    \begin{minipage}[t]{.485\linewidth}
    \centering
    \includegraphics[width=\linewidth,height=.215\textheight,keepaspectratio]{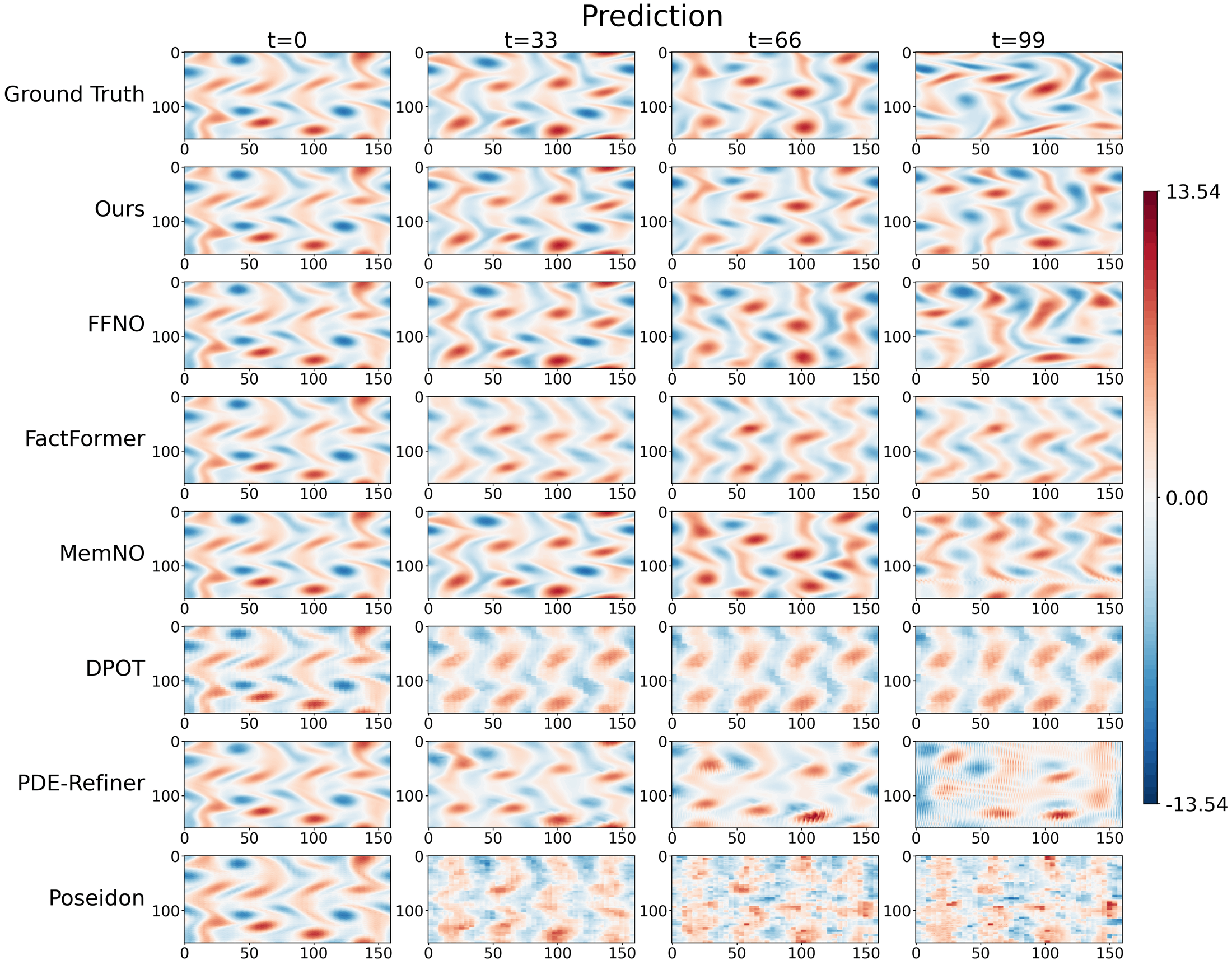}
    \captionof{figure}{Kolmogorov flow predictions}
    \label{fig:kolm_prediction}
  \end{minipage}\hfill
  \begin{minipage}[t]{.485\linewidth}
    \centering
    \includegraphics[width=\linewidth,height=.215\textheight,keepaspectratio]{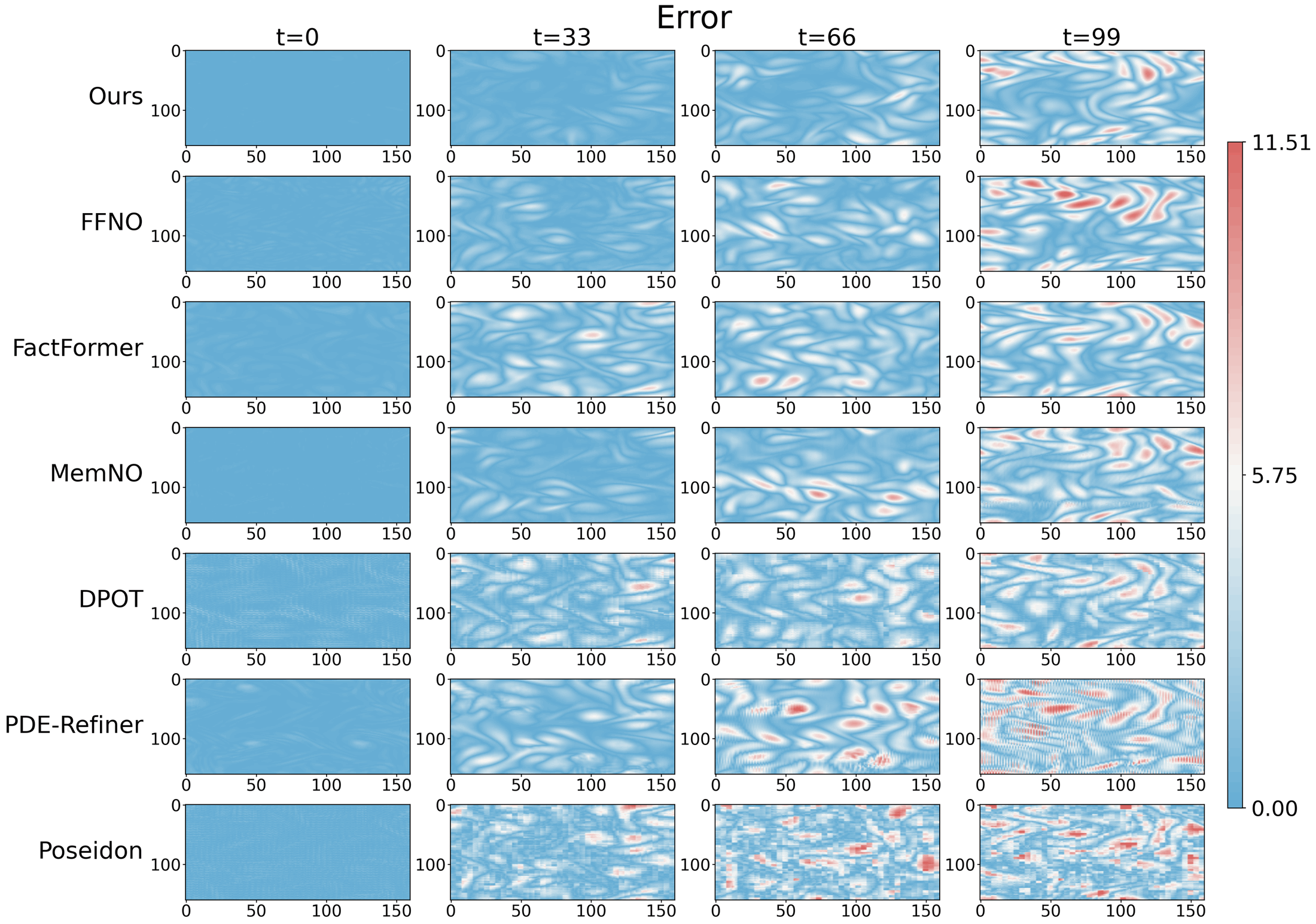}
    \captionof{figure}{Kolmogorov flow spatial errors}
    \label{fig:kolm_error}
  \end{minipage}\par
    \begin{minipage}[t]{.485\linewidth}
    \centering
    \includegraphics[width=\linewidth,height=.215\textheight,keepaspectratio]{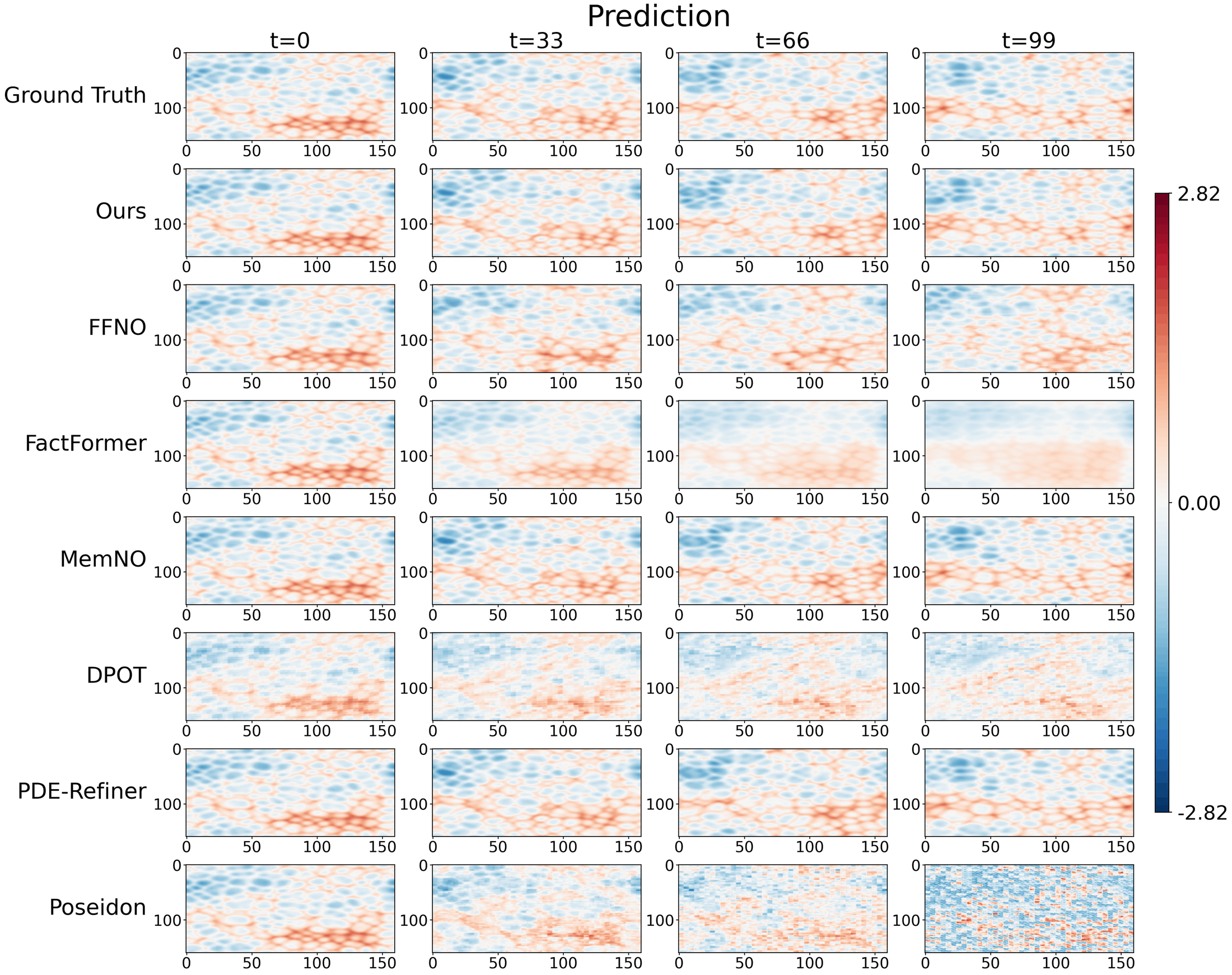}
    \captionof{figure}{KS predictions}
    \label{fig:ks_prediction}
  \end{minipage}\hfill
  \begin{minipage}[t]{.485\linewidth}
    \centering
    \includegraphics[width=\linewidth,height=.215\textheight,keepaspectratio]{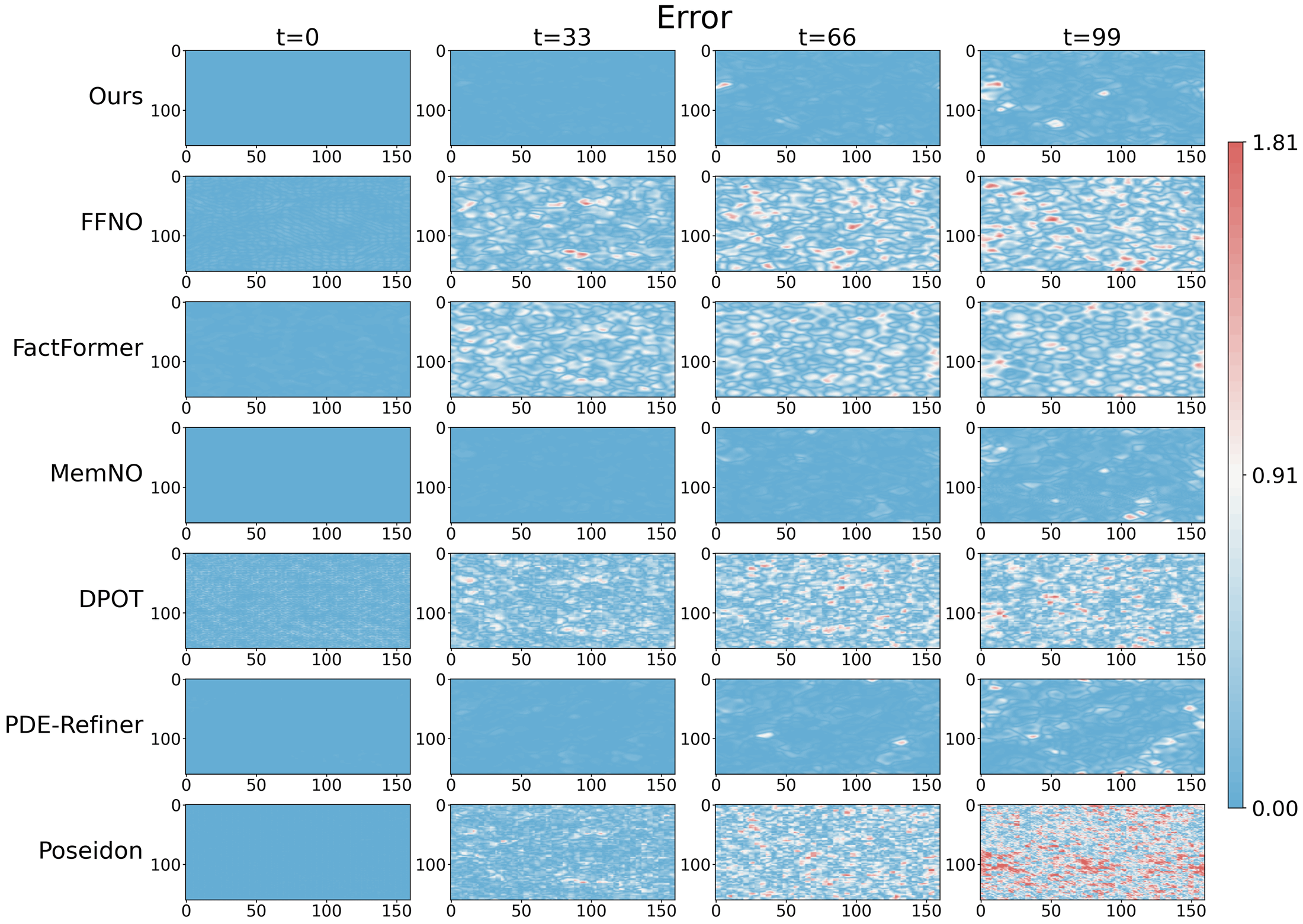}
    \captionof{figure}{KS spatial errors}
    \label{fig:ks_error}
  \end{minipage}\par
\end{minipage}
\par
\noindent\begin{minipage}{\linewidth}
\begin{minipage}[t]{.485\linewidth}
    \centering
    \includegraphics[width=\linewidth,height=.215\textheight,keepaspectratio]{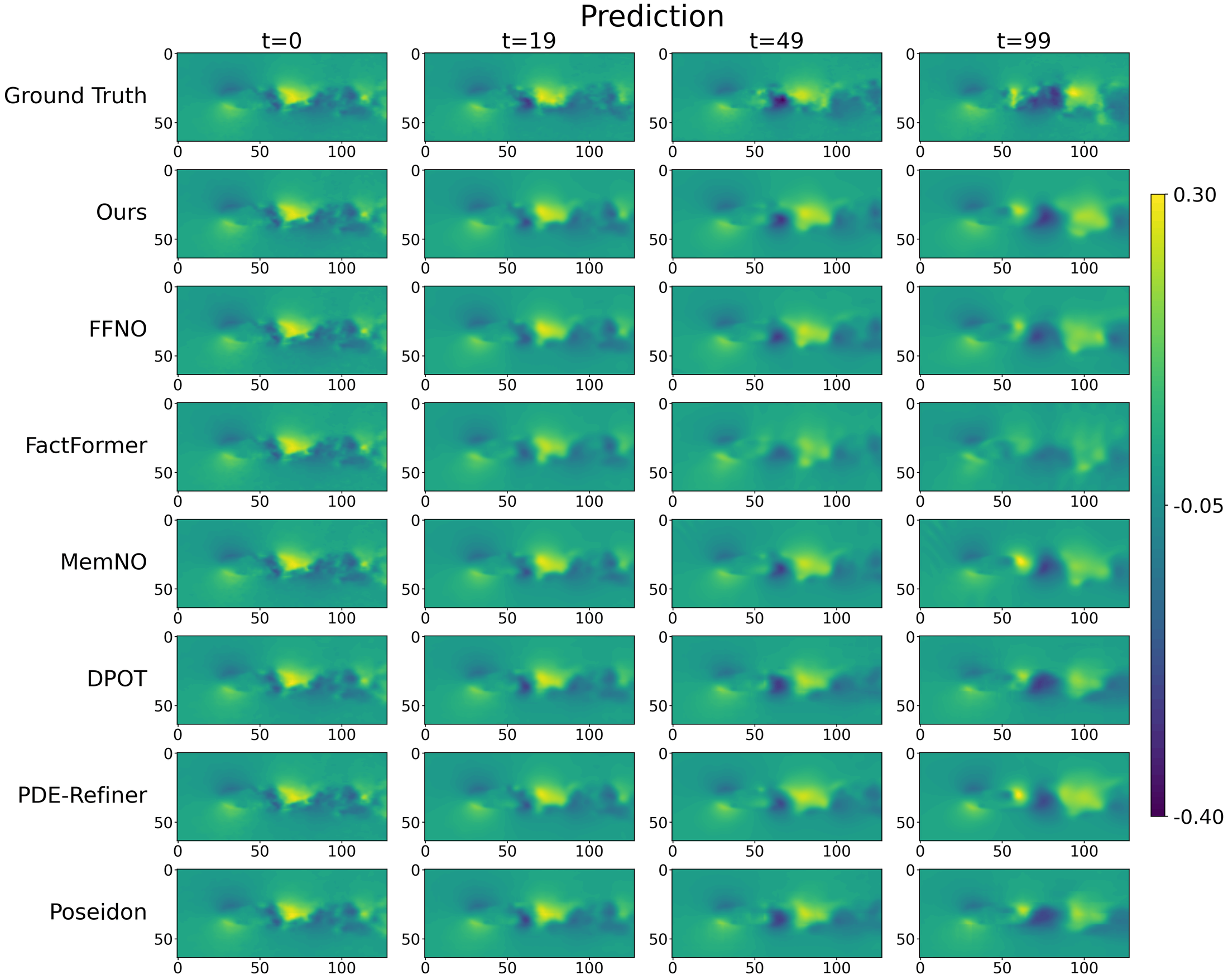}
    \captionof{figure}{Cylinder $v$ predictions}
    \label{fig:cylinder_v_prediction}
  \end{minipage}\hfill
  \begin{minipage}[t]{.485\linewidth}
    \centering
    \includegraphics[width=\linewidth,height=.215\textheight,keepaspectratio]{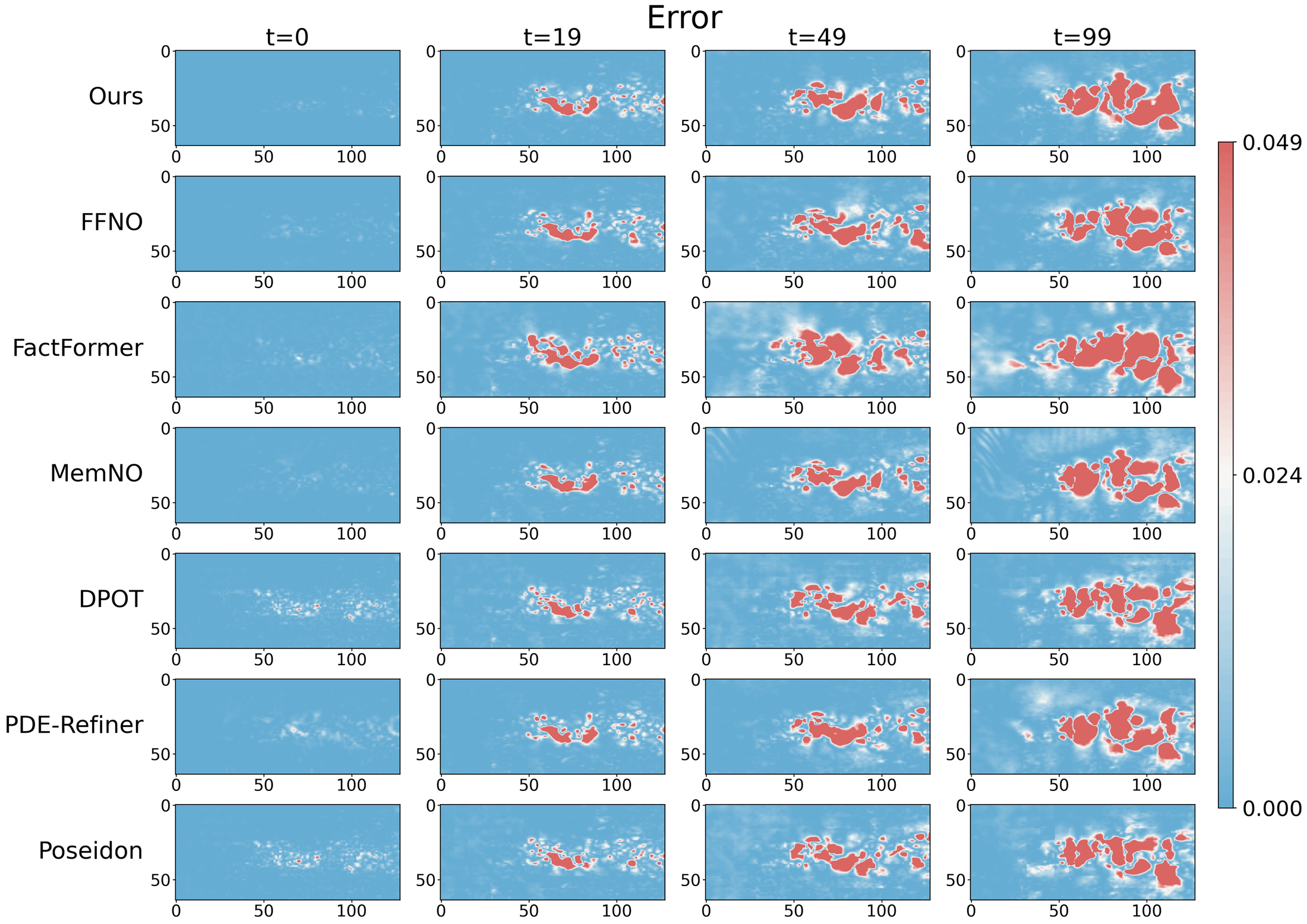}
    \captionof{figure}{Cylinder $v$ spatial errors}
    \label{fig:cylinder_error}
  \end{minipage}\par
    \begin{minipage}[t]{.485\linewidth}
    \centering
    \includegraphics[width=\linewidth,height=.215\textheight,keepaspectratio]{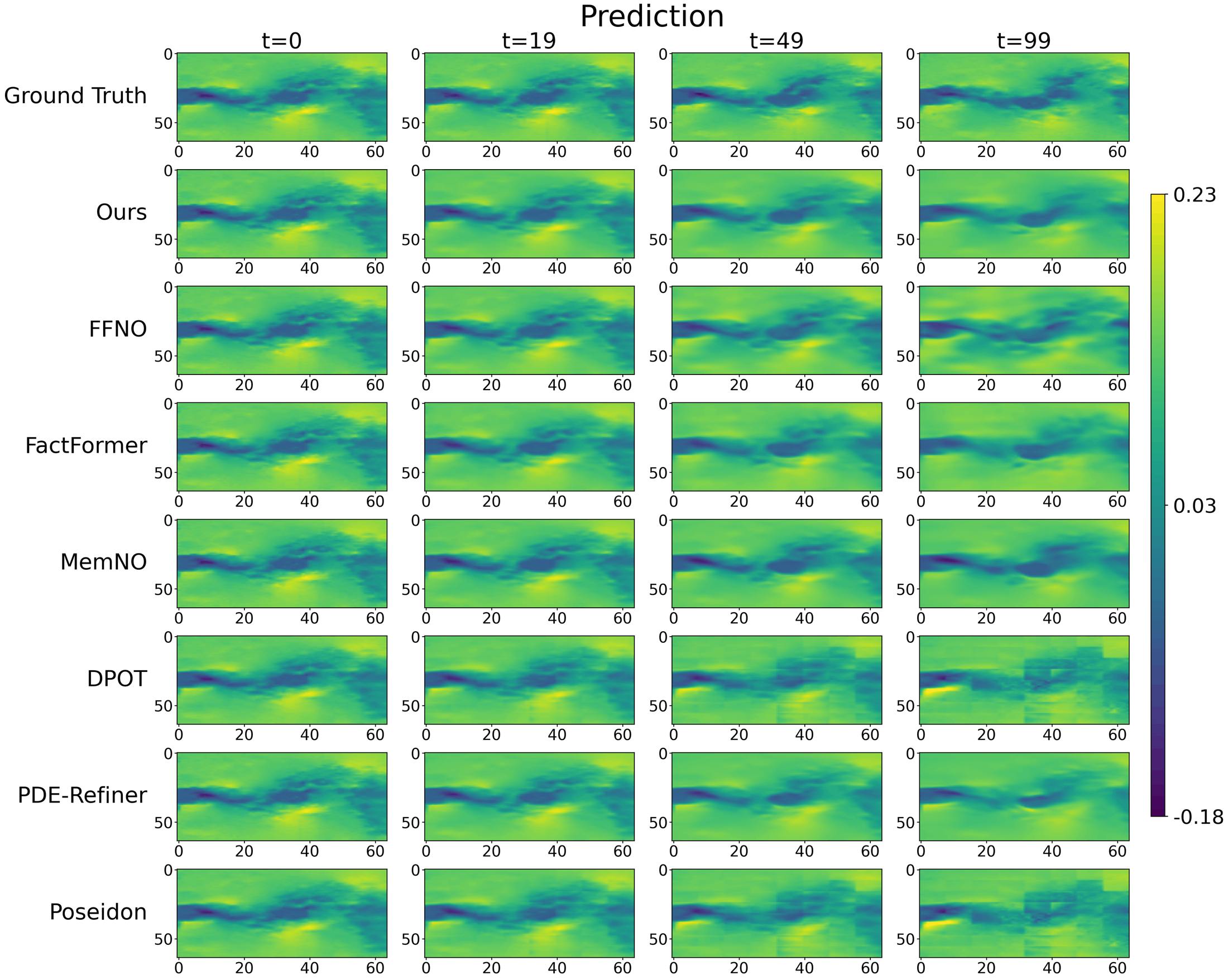}
    \captionof{figure}{FSI $u$ predictions}
    \label{fig:fsi_u_prediction}
  \end{minipage}\hfill
  \begin{minipage}[t]{.485\linewidth}
    \centering
    \includegraphics[width=\linewidth,height=.215\textheight,keepaspectratio]{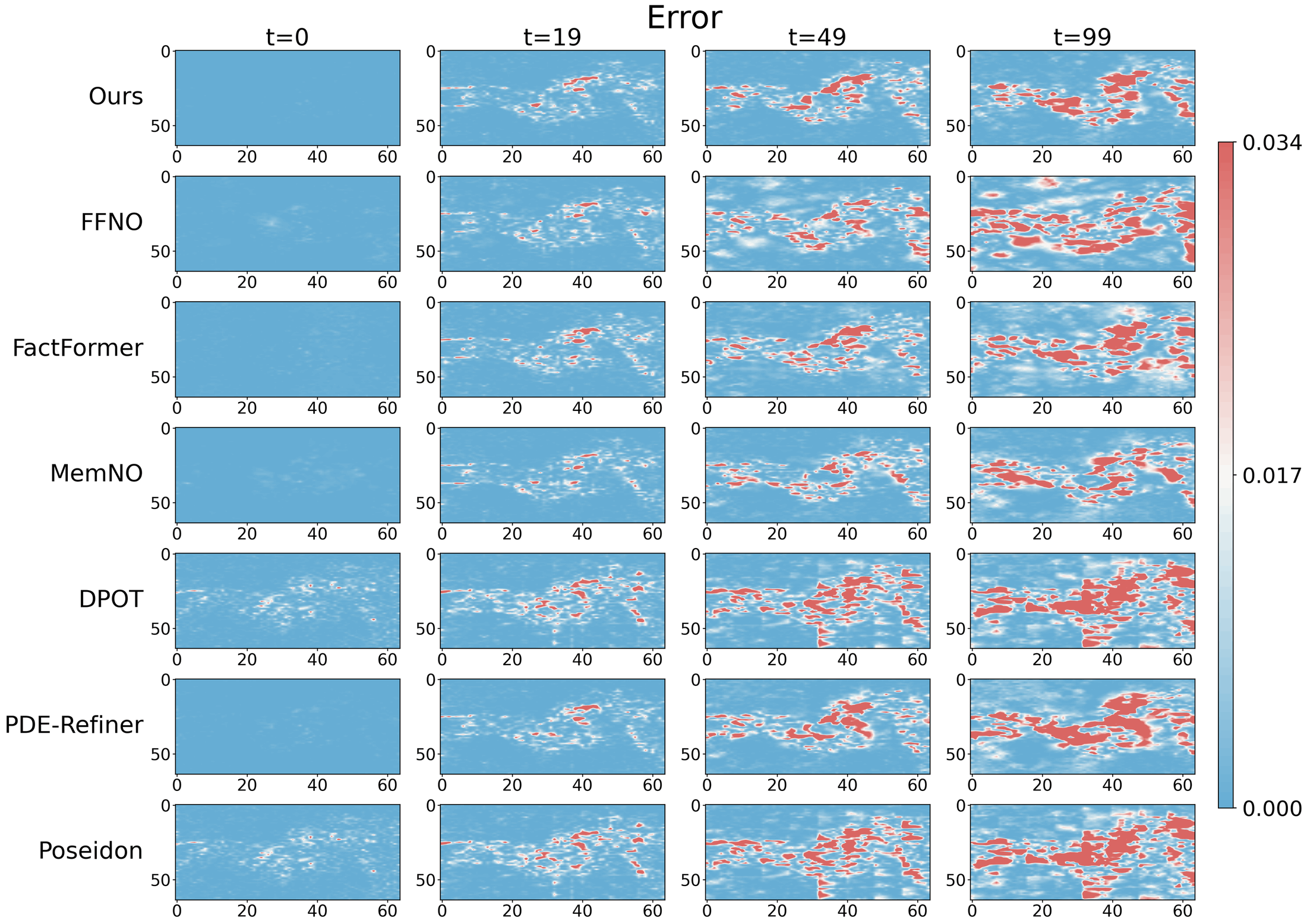}
    \captionof{figure}{FSI $u$ spatial errors}
    \label{fig:fsi_u_error}
  \end{minipage}\par
    \begin{minipage}[t]{.485\linewidth}
    \centering
    \includegraphics[width=\linewidth,height=.215\textheight,keepaspectratio]{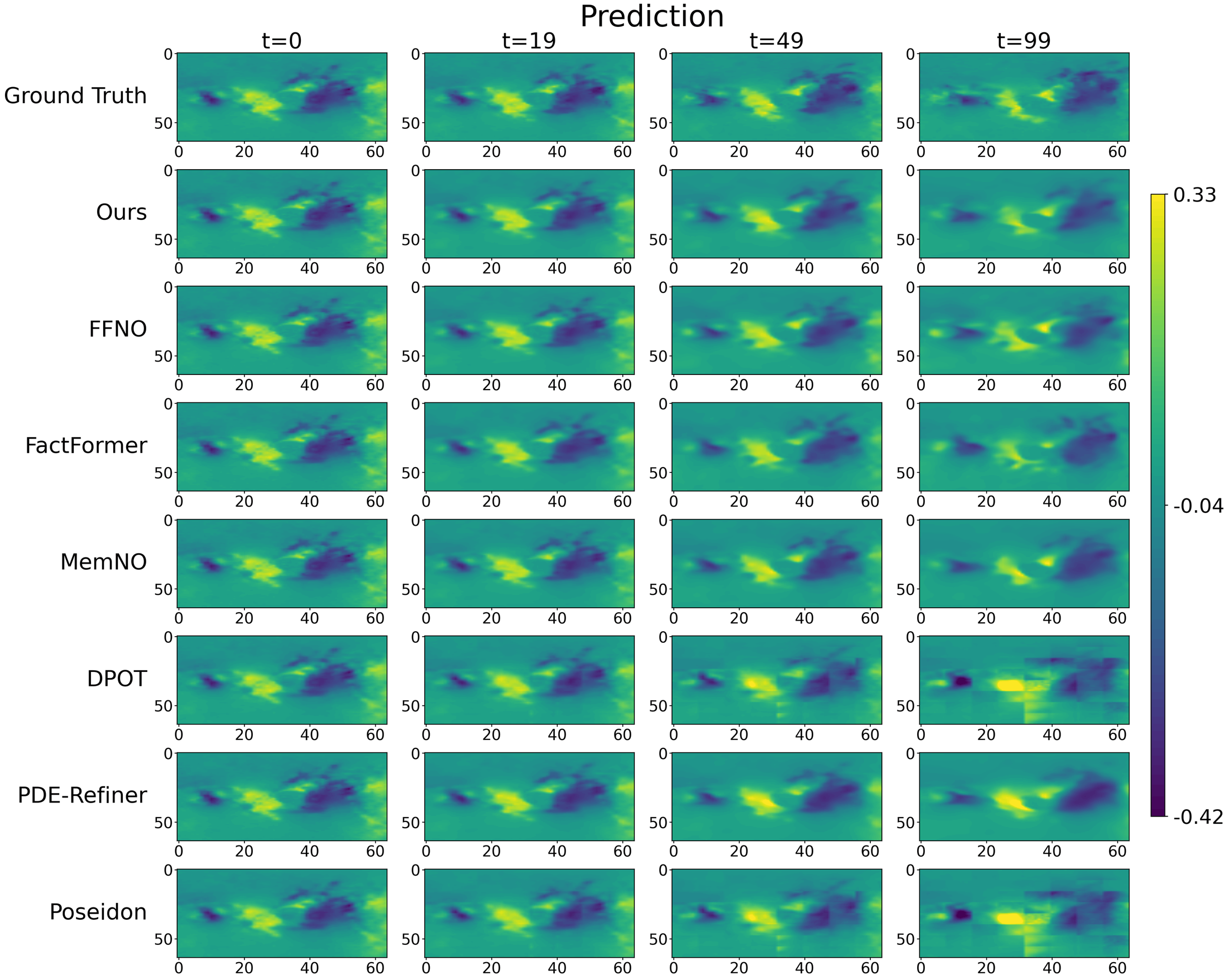}
    \captionof{figure}{FSI $v$ predictions}
    \label{fig:fsi_v_prediction}
  \end{minipage}\hfill
  \begin{minipage}[t]{.485\linewidth}
    \centering
    \includegraphics[width=\linewidth,height=.215\textheight,keepaspectratio]{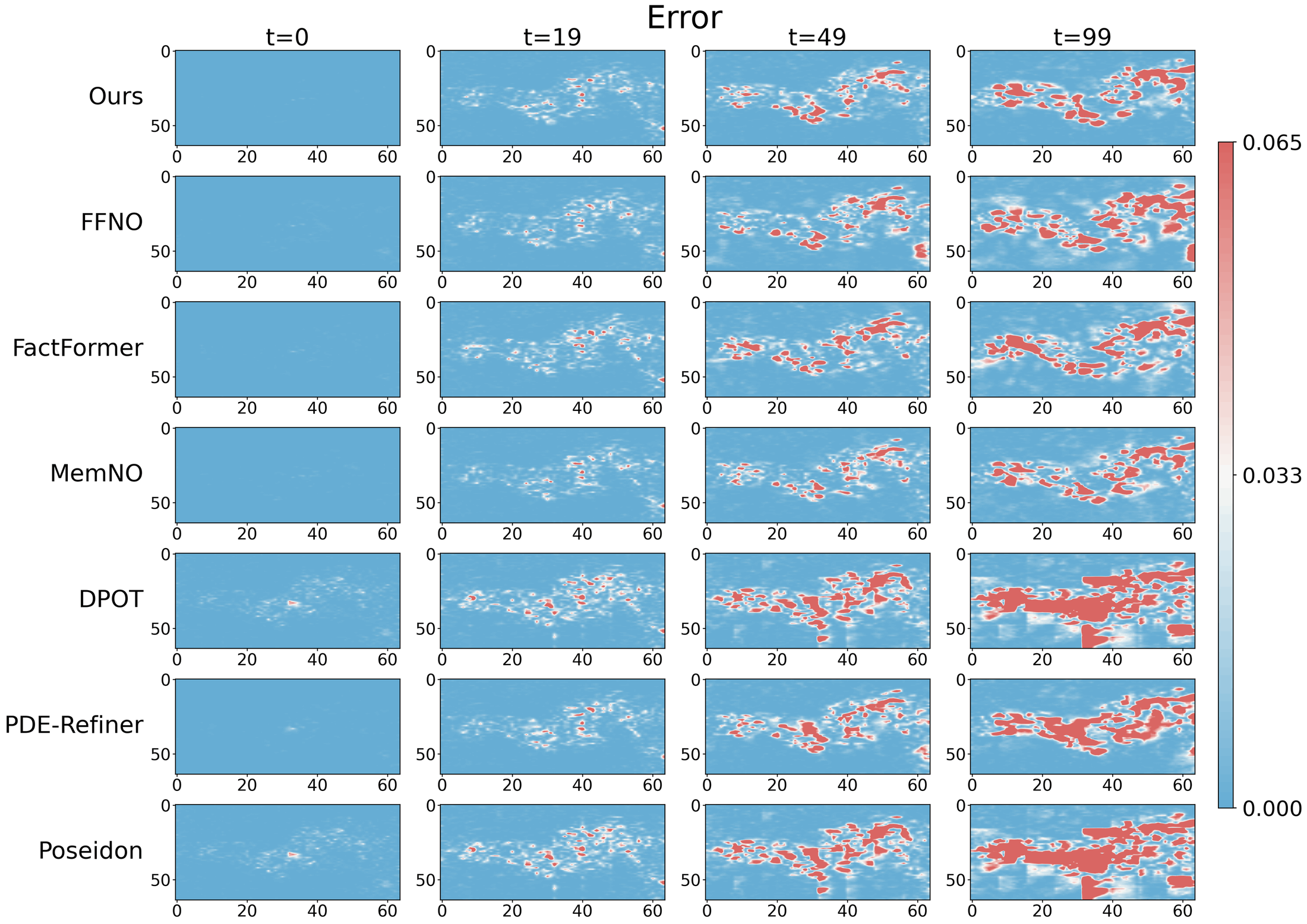}
    \captionof{figure}{FSI $v$ spatial errors}
    \label{fig:fsi_v_error}
  \end{minipage}\par
\end{minipage}
\par
\noindent\begin{minipage}{\linewidth}
\centering
\footnotesize
\setlength{\tabcolsep}{5pt}
\renewcommand{\arraystretch}{1.13}
\captionof{table}{Mean paired reduction in final-step nRMSE from
restoring the commutator signal. Positive values indicate improvement.}
\label{tab:module3_decomposed_gain}
\sbox0{%
\begin{tabular}{lccccc}
\toprule
System & Burgers & KS & Kolmogorov & Cylinder & FSI \\
\midrule
Terminal gain $\uparrow$
& 0.0028 & 0.0098 & 0.0266 & 0.0697 & 0.0053 \\
\bottomrule
\end{tabular}
}
\ifdim\wd0>\linewidth
  \resizebox{\linewidth}{!}{\usebox0}%
\else
  \usebox0
\fi
\end{minipage}
\par
\noindent\begin{minipage}{\linewidth}
\centering
\footnotesize
\setlength{\tabcolsep}{4pt}
\renewcommand{\arraystretch}{1.13}
\captionof{table}{Temporal growth exponents fitted over
$k\in\{1,2,4,8,16\}$. Burgers and KS use step units; the other
systems use the listed physical time increments. $N_{\mathrm{train}}$
denotes the training-trajectory count.}
\label{tab:module3_time_scale_diagnostics}
\sbox0{%
\begin{tabular}{lccccc}
\toprule
System
& Time increment
& $N_{\mathrm{train}}$
& \shortstack{State commutator\\slope}
& \shortstack{State increment\\slope}
& \shortstack{Normalized evidence\\slope} \\
\midrule
Burgers
& Step & 30 & 0.8927 & 0.9126 & 0.8511 \\
KS
& Step & 30 & 0.9528 & 0.9579 & 0.8247 \\
Kolmogorov
& $0.1$ & 30 & 0.7350 & 0.7847 & 0.4079 \\
Cylinder
& $0.005$ & 84 & 1.0473 & 0.9667 & 0.7003 \\
FSI
& $0.002$ & 42 & 1.0080 & 0.9967 & 0.7937 \\
\bottomrule
\end{tabular}
}
\ifdim\wd0>\linewidth
  \resizebox{\linewidth}{!}{\usebox0}%
\else
  \usebox0
\fi
\end{minipage}
\par
\noindent\begin{minipage}{\linewidth}
\centering
\footnotesize
\setlength{\tabcolsep}{4.5pt}
\renewcommand{\arraystretch}{1.13}
\captionof{table}{Mean stepwise relative $\ell_2$ error over frame-by-frame
autoregressive rollouts under inference-time signal scaling.
$\lambda=0$ removes the signal; $\lambda=1$ retains its default amplitude
in this recursive protocol. Boldface marks the lowest error in each row.}
\label{tab:module3_lambda_sensitivity}
\sbox0{%
\begin{tabular}{lcccccc}
\toprule
System
& $\lambda=0$
& $\lambda=0.25$
& $\lambda=0.5$
& $\lambda=1$
& $\lambda=2$
& $\lambda=4$ \\
\midrule
Burgers
& 0.0589
& 0.0542
& \textbf{0.0540}
& 0.0544
& 0.0585
& 0.0711 \\
KS
& 0.5692
& 0.5636
& 0.5612
& \textbf{0.5591}
& 0.5748
& 0.5920 \\
Kolmogorov
& 0.5689
& 0.5559
& \textbf{0.5512}
& \textbf{0.5512}
& 0.5796
& 0.5920 \\
Cylinder
& 0.1942
& 0.1683
& 0.1534
& \textbf{0.1455}
& 0.1504
& 0.1700 \\
FSI
& 0.2304
& 0.2202
& \textbf{0.2178}
& 0.2220
& 0.2499
& 0.3622 \\
\bottomrule
\end{tabular}
}
\ifdim\wd0>\linewidth
  \resizebox{\linewidth}{!}{\usebox0}%
\else
  \usebox0
\fi
\end{minipage}
\par
\noindent\begin{minipage}{\linewidth}
\centering
\footnotesize
\setlength{\tabcolsep}{3.2pt}
\renewcommand{\arraystretch}{1.13}
\captionof{table}{Burgers under $4\times$ spatial coarsening. Metrics follow the main-text definitions; boldface marks the best mean in each column.}
\label{tab:main_comparison_burgers_4x}
\sbox0{%
\begin{tabular}{lccccc}
\toprule
Model
& Rel. $\ell_2$ $\downarrow$
& Traj. nRMSE $\downarrow$
& Final nRMSE $\downarrow$
& HF Error $\downarrow$
& Mean Corr. $\uparrow$ \\
\midrule
DPOT
& $1.8691$
& $1.6318$
& $5.6546$
& $4.5972$
& $0.6446$ \\
FactFormer
& $0.1022$
& $0.1108$
& $0.1730$
& $0.2621$
& $0.9932$ \\
F-FNO
& $0.1598$
& $0.1368$
& $0.6986$
& $0.1501$
& $0.9782$ \\
MemNO
& $0.0921$
& $0.0850$
& $0.3645$
& $0.1279$
& $0.9925$ \\
PDE-Refiner
& $0.2297$
& $0.1868$
& $0.9345$
& $0.1825$
& $0.9632$ \\
Poseidon
& $1.9680$
& $1.7388$
& $5.3749$
& $4.9805$
& $0.6223$ \\
\midrule
\textbf{CoMNO}
& $\mathbf{0.0369}$
& $\mathbf{0.0397}$
& $\mathbf{0.0729}$
& $\mathbf{0.0968}$
& $\mathbf{0.9991}$ \\
\bottomrule
\end{tabular}
}
\ifdim\wd0>\linewidth
  \resizebox{\linewidth}{!}{\usebox0}%
\else
  \usebox0
\fi
\end{minipage}
\par
\noindent\begin{minipage}{\linewidth}
\centering
\footnotesize
\setlength{\tabcolsep}{3.2pt}
\renewcommand{\arraystretch}{1.13}
\captionof{table}{KS under $4\times$ spatial coarsening. Metrics follow the main-text definitions; boldface marks the best mean in each column.}
\label{tab:main_comparison_ks_4x}
\sbox0{%
\begin{tabular}{lccccc}
\toprule
Model
& Rel. $\ell_2$ $\downarrow$
& Traj. nRMSE $\downarrow$
& Final nRMSE $\downarrow$
& HF Error $\downarrow$
& Mean Corr. $\uparrow$ \\
\midrule
DPOT
& $1.0642$
& $1.0178$
& $1.4467$
& $1.1944$
& $0.8837$ \\
FactFormer
& $0.2644$
& $0.2072$
& $0.5552$
& $0.2652$
& $0.9643$ \\
F-FNO
& $0.4195$
& $0.3396$
& $0.8282$
& $0.4248$
& $0.9107$ \\
MemNO
& $0.2727$
& $0.2146$
& $0.5772$
& $0.2722$
& $0.9620$ \\
PDE-Refiner
& $0.3101$
& $0.2456$
& $0.6352$
& $0.3124$
& $0.9508$ \\
Poseidon
& $1.0137$
& $0.9754$
& $1.2768$
& $1.1369$
& $0.8903$ \\
\midrule
\textbf{CoMNO}
& $\mathbf{0.2539}$
& $\mathbf{0.2066}$
& $\mathbf{0.5531}$
& $\mathbf{0.2587}$
& $\mathbf{0.9662}$ \\
\bottomrule
\end{tabular}
}
\ifdim\wd0>\linewidth
  \resizebox{\linewidth}{!}{\usebox0}%
\else
  \usebox0
\fi
\end{minipage}
\par
\noindent\begin{minipage}{\linewidth}
\centering
\footnotesize
\setlength{\tabcolsep}{3.2pt}
\renewcommand{\arraystretch}{1.13}
\captionof{table}{Kolmogorov flow under $4\times$ spatial coarsening. Metrics follow the main-text definitions; boldface marks the best mean in each column.}
\label{tab:main_comparison_kolm_4x}
\sbox0{%
\begin{tabular}{lccccc}
\toprule
Model
& Rel. $\ell_2$ $\downarrow$
& Traj. nRMSE $\downarrow$
& Final nRMSE $\downarrow$
& HF Error $\downarrow$
& Mean Corr. $\uparrow$ \\
\midrule
DPOT
& $0.8372$
& $0.8274$
& $0.9103$
& $1.1376$
& $0.5324$ \\
FactFormer
& $0.8107$
& $0.7500$
& $1.1085$
& $1.0817$
& $0.6616$ \\
F-FNO
& $0.7661$
& $0.7030$
& $1.1302$
& $1.0826$
& $0.7068$ \\
MemNO
& $0.6828$
& $0.6060$
& $1.0607$
& $0.9691$
& $0.7650$ \\
PDE-Refiner
& $0.8022$
& $0.7245$
& $1.1724$
& $1.0424$
& $0.6799$ \\
Poseidon
& $0.8341$
& $0.8235$
& $0.9203$
& $1.1445$
& $0.5513$ \\
\midrule
\textbf{CoMNO}
& $\mathbf{0.5523}$
& $\mathbf{0.4777}$
& $\mathbf{0.9048}$
& $\mathbf{0.8291}$
& $\mathbf{0.8340}$ \\
\bottomrule
\end{tabular}
}
\ifdim\wd0>\linewidth
  \resizebox{\linewidth}{!}{\usebox0}%
\else
  \usebox0
\fi
\end{minipage}
\par
\noindent\begin{minipage}{\linewidth}
\centering
\footnotesize
\setlength{\tabcolsep}{3.2pt}
\renewcommand{\arraystretch}{1.13}
\captionof{table}{Burgers under $8\times$ spatial coarsening. Metrics follow the main-text definitions; boldface marks the best mean in each column.}
\label{tab:main_comparison_burgers_8x}
\sbox0{%
\begin{tabular}{lccccc}
\toprule
Model
& Rel. $\ell_2$ $\downarrow$
& Traj. nRMSE $\downarrow$
& Final nRMSE $\downarrow$
& HF Error $\downarrow$
& Mean Corr. $\uparrow$ \\
\midrule
DPOT
& $0.5241$
& $0.5623$
& $0.8472$
& $0.7066$
& $0.7943$ \\
FactFormer
& $0.0791$
& $0.0819$
& $0.2175$
& $\mathbf{0.0814}$
& $0.9959$ \\
F-FNO
& $0.2612$
& $0.2406$
& $0.8523$
& $0.2071$
& $0.9495$ \\
MemNO
& $0.1186$
& $0.1177$
& $0.3995$
& $0.1198$
& $0.9885$ \\
PDE-Refiner
& $0.2614$
& $0.2247$
& $1.0682$
& $0.1300$
& $0.9555$ \\
Poseidon
& $0.5251$
& $0.5634$
& $0.8497$
& $0.7071$
& $0.7935$ \\
\midrule
\textbf{CoMNO}
& $\mathbf{0.0733}$
& $\mathbf{0.0785}$
& $\mathbf{0.1216}$
& $0.0999$
& $\mathbf{0.9965}$ \\
\bottomrule
\end{tabular}
}
\ifdim\wd0>\linewidth
  \resizebox{\linewidth}{!}{\usebox0}%
\else
  \usebox0
\fi
\end{minipage}
\par
\noindent\begin{minipage}{\linewidth}
\centering
\footnotesize
\setlength{\tabcolsep}{3.2pt}
\renewcommand{\arraystretch}{1.13}
\captionof{table}{KS under $8\times$ spatial coarsening. Metrics follow the main-text definitions; boldface marks the best mean in each column.}
\label{tab:main_comparison_ks_8x}
\sbox0{%
\begin{tabular}{lccccc}
\toprule
Model
& Rel. $\ell_2$ $\downarrow$
& Traj. nRMSE $\downarrow$
& Final nRMSE $\downarrow$
& HF Error $\downarrow$
& Mean Corr. $\uparrow$ \\
\midrule
DPOT
& $1.0016$
& $0.9849$
& $1.1943$
& $1.2075$
& $0.4025$ \\
FactFormer
& $0.8185$
& $0.7359$
& $1.1958$
& $0.8782$
& $0.6521$ \\
F-FNO
& $1.3084$
& $1.1819$
& $1.9977$
& $1.2119$
& $0.4421$ \\
MemNO
& $1.3343$
& $1.1852$
& $1.7395$
& $1.5925$
& $0.3159$ \\
PDE-Refiner
& $0.9359$
& $0.8854$
& $1.1956$
& $1.0869$
& $0.5288$ \\
Poseidon
& $0.9971$
& $0.9812$
& $1.1916$
& $1.1881$
& $0.4375$ \\
\midrule
\textbf{CoMNO}
& $\mathbf{0.7911}$
& $\mathbf{0.7314}$
& $\mathbf{0.9894}$
& $\mathbf{0.8120}$
& $\mathbf{0.6924}$ \\
\bottomrule
\end{tabular}
}
\ifdim\wd0>\linewidth
  \resizebox{\linewidth}{!}{\usebox0}%
\else
  \usebox0
\fi
\end{minipage}
\par
\noindent\begin{minipage}{\linewidth}
\centering
\footnotesize
\setlength{\tabcolsep}{3.2pt}
\renewcommand{\arraystretch}{1.13}
\captionof{table}{Kolmogorov flow under $8\times$ spatial coarsening. Metrics follow the main-text definitions; boldface marks the best mean in each column.}
\label{tab:main_comparison_kolm_8x}
\sbox0{%
\begin{tabular}{lccccc}
\toprule
Model
& Rel. $\ell_2$ $\downarrow$
& Traj. nRMSE $\downarrow$
& Final nRMSE $\downarrow$
& HF Error $\downarrow$
& Mean Corr. $\uparrow$ \\
\midrule
DPOT
& $0.9186$
& $0.9099$
& $1.0329$
& $0.8337$
& $0.5181$ \\
FactFormer
& $0.6587$
& $0.5774$
& $1.0420$
& $0.5419$
& $0.7809$ \\
F-FNO
& $0.7432$
& $0.6496$
& $1.1729$
& $0.5667$
& $0.7510$ \\
MemNO
& $0.6075$
& $0.5299$
& $1.0175$
& $0.4965$
& $0.8149$ \\
PDE-Refiner
& $0.6076$
& $0.5251$
& $1.0547$
& $0.4997$
& $0.8146$ \\
Poseidon
& $0.9390$
& $0.9298$
& $1.0765$
& $0.8403$
& $0.5031$ \\
\midrule
\textbf{CoMNO}
& $\mathbf{0.6012}$
& $\mathbf{0.5235}$
& $\mathbf{0.9765}$
& $\mathbf{0.4919}$
& $\mathbf{0.8199}$ \\
\bottomrule
\end{tabular}
}
\ifdim\wd0>\linewidth
  \resizebox{\linewidth}{!}{\usebox0}%
\else
  \usebox0
\fi
\end{minipage}
\par

\section{Supplementary Results for the Main Experiments}
\label{app:main_experiments}

\subsection{Main Comparison}
\label{sec:main_supp}

\paragraph{Prediction across systems.}
Tables~\ref{tab:burgers_additional_metrics}--\ref{tab:main_comparison_fsi}
reproduce the five-system main comparison under the evaluation protocol
specified in the main text. Horizon-specific recursive
extensions are reported separately in Appendix~\ref{sec:rollout_supp}.
CoMNO leads 24 of 25 comparisons, including field, trajectory, and terminal
errors and spatial correlation on all five systems. KS shows a joint improvement
in rollout accuracy and fine-scale prediction. On Kolmogorov flow, the lower errors
accompany better spatial agreement. FSI shows smaller gains in
trajectory and terminal accuracy. Agreement across full-trajectory and
stepwise errors connects the improvement to both the complete
forecast and predicted states. Strong HF fidelity and
higher correlation connect these gains to the spatial
structure of the predicted fields.

\subsection{Functional Role of the Commutator Signal}
\label{app:module3_commutator_diagnostics}

We extend the frozen-checkpoint intervention in
Section~\ref{subsec:commutator_diagnostics} with terminal prediction,
temporal growth, and amplitude sensitivity diagnostics.
For the prediction interventions, all learned parameters remain
fixed, and $\bar e_t\mapsto\lambda\bar e_t$ is applied during
inference. Evidence is recomputed along each resulting trajectory
using the same schedule as the unmodified predictor.
The interventions use the main-text test cases.
Errors are first summarized within each trajectory;
windows from the same physical recording are averaged within
their trajectory group before averaging across independent groups.
Paired gains compare the same evaluation units with and without
the commutator signal.

\paragraph{Terminal prediction.}
Table~\ref{tab:module3_decomposed_gain} applies the main-text paired
gain definition to final-step nRMSE. Each entry is the mean error
with the signal removed minus that with the signal restored.
All five gains are positive, with the largest absolute reduction
on Cylinder. The benefit therefore extends to the final predicted
state, complementing the full-rollout and late-horizon results
in the main text.

\paragraph{Temporal growth of the evidence.}
Starting from matched field and memory states, we repeatedly apply
the shared provisional transition with zero current evidence on
the observation grid and its coarsened versions. At each horizon,
we compare evolution followed by downsampling with downsampling
followed by evolution. The comparison uses the same normalization,
lifting, and evidence construction as Operator Commutator in
Section~\ref{sec:commutator}.
For each diagnostic magnitude $m$, we fit
\begin{equation}
\log m(k\Delta t)
=
a\log(k\Delta t)+b,
\label{eq:commutator_temporal_growth}
\end{equation}
by least squares over $k\in\{1,2,4,8,16\}$.
Here, $\Delta t$ is the time increment per provisional step,
$a$ is the fitted growth exponent, and $b$ is the intercept.
Table~\ref{tab:module3_time_scale_diagnostics} reports $a$ for
the raw state commutator, state increment, and normalized evidence.
Burgers and KS use step units; the remaining systems use their
listed physical time increments.
The raw state commutator and state increment have similar fitted
exponents. Normalized evidence has a positive exponent smaller
than both on every system, indicating slower temporal growth
while retaining sensitivity to the evolving trajectory over
the tested horizons.

\paragraph{Sensitivity to signal amplitude.}
Table~\ref{tab:module3_lambda_sensitivity} feeds predictions back at each
step, unlike the single-sequence forecasts in Table~\ref{tab:main_comparison_ks}.
Stepwise relative $\ell_2$ errors are averaged over the rollout and then
over independent evaluation units. The default amplitude achieves the lowest
displayed error on KS and Cylinder and ties with half amplitude
on Kolmogorov flow. Half amplitude gives the lowest displayed
errors on Burgers and FSI.
All three tested amplitudes between $0.25$ and $1$ improve over
signal removal on every system. Increasing the amplitude beyond
$1$ raises error relative to the default on all five systems,
although Cylinder remains better than signal removal at every
tested nonzero amplitude.
These results support the predictive value of the commutator
signal and show that its contribution depends on amplitude,
with moderate scaling providing the most consistent gains.

\subsection{Performance under Spatial Coarsening}
\label{app:coarse_observation}

\paragraph{Protocol and metrics.}
We coarsen the observed histories of Burgers, KS, and Kolmogorov
flow by factors of $4\times$ and $8\times$, while retaining future
reference fields on the original grid. Each method is trained
separately for each factor using the same data splits, temporal
windows, normalization procedures, and evaluation protocol.
Tables~\ref{tab:main_comparison_burgers_4x}
to~\ref{tab:main_comparison_kolm_8x} report five prediction metrics
using the main-text definitions. Entries report means. In Table~\ref{tab:coarse_gain},
Mean nRMSE and Spectral Error correspond to Traj.\ nRMSE and
HF Error, respectively. For each setting and metric, the relative
error change is computed from the reported means as
\begin{equation}
\Delta_m
=
100\left(
\frac{\overline{E}_{\mathrm{CoMNO},m}}
{\min_{b\in\mathcal{B}}\overline{E}_{b,m}}-1
\right),
\label{eq:coarsening_relative_change}
\end{equation}
where $\mathcal{B}$ contains the six baselines and $m$ indexes
one of the three summarized error metrics. The best baseline
is selected independently for each setting and metric;
negative values indicate improvement.

\paragraph{Results.}
CoMNO achieves the lowest relative $\ell_2$ error in all six
settings. Across the three metrics summarized in
Table~\ref{tab:coarse_gain}, it improves 17 of 18 comparisons,
with gains spanning all three systems and multiple metrics at both coarsening factors.
On Burgers, CoMNO leads trajectory and terminal accuracy at both
factors and spectral accuracy at $4\times$, while FactFormer
leads spectral accuracy at $8\times$. On KS, CoMNO leads all three
summarized metrics at both factors, with larger terminal and spectral
gains at $8\times$. On Kolmogorov flow, CoMNO leads all three
metrics at both factors, with larger gains in trajectory and spectral
accuracy at $4\times$ and larger gains in terminal accuracy
at $8\times$. These results show that CoMNO's predictive benefits persist
with reduced spatial information, with gains distributed
across trajectory, terminal, and spectral fidelity.

\par
\noindent\begin{minipage}{\linewidth}
\centering
\footnotesize
\setlength{\tabcolsep}{3.2pt}
\renewcommand{\arraystretch}{1.13}
\captionof{table}{Rollout accuracy on Burgers. Relative $\ell_2$ uses the first $H$ predictions; Final Step nRMSE uses step $197$.}
\label{tab:rollout_burgers}
\sbox0{%
\begin{tabular}{lccccc}
\toprule
Model
& \multicolumn{4}{c}{Rel. $\ell_2$ $\downarrow$}
& \shortstack{Final Step\\nRMSE} $\downarrow$ \tabularnewline
\cmidrule(lr){2-5}
\cmidrule(lr){6-6}
& $H=20$
& $H=50$
& $H=100$
& $H=197$
& $H=197$ \tabularnewline
\midrule
DPOT
& $0.1072$
& $0.1712$
& $0.3679$
& $0.6317$
& $0.9955$ \tabularnewline
FactFormer
& $0.0461$
& $0.0783$
& $0.1063$
& $0.1557$
& $0.3433$ \tabularnewline
F-FNO
& $0.0725$
& $0.1506$
& $0.4601$
& $1.3937$
& $3.6467$ \tabularnewline
MemNO
& $0.0442$
& $0.1082$
& $0.2011$
& $0.3494$
& $0.7902$ \tabularnewline
PDE-Refiner
& $0.1017$
& $0.1791$
& $0.3249$
& $0.7272$
& $1.7102$ \tabularnewline
Poseidon
& $0.1264$
& $0.2386$
& $0.4233$
& $0.7154$
& $1.2365$ \tabularnewline
\textbf{CoMNO}
& $\mathbf{0.0089}$
& $\mathbf{0.0235}$
& $\mathbf{0.0549}$
& $\mathbf{0.1315}$
& $\mathbf{0.2812}$ \tabularnewline
\bottomrule
\end{tabular}
}
\ifdim\wd0>\linewidth
  \resizebox{\linewidth}{!}{\usebox0}%
\else
  \usebox0
\fi
\end{minipage}
\par
\noindent\begin{minipage}{\linewidth}
\centering
\footnotesize
\setlength{\tabcolsep}{3.2pt}
\renewcommand{\arraystretch}{1.13}
\captionof{table}{Rollout accuracy on KS. Relative $\ell_2$ uses the first $H$ predictions; Final Step nRMSE uses step $197$.}
\label{tab:rollout_ks}
\sbox0{%
\begin{tabular}{lccccc}
\toprule
Model
& \multicolumn{4}{c}{Rel. $\ell_2$ $\downarrow$}
& \shortstack{Final Step\\nRMSE} $\downarrow$ \tabularnewline
\cmidrule(lr){2-5}
\cmidrule(lr){6-6}
& $H=20$
& $H=50$
& $H=100$
& $H=197$
& $H=197$ \tabularnewline
\midrule
DPOT
& $0.1159$
& $0.4763$
& $0.8799$
& $1.1432$
& $1.3979$ \tabularnewline
FactFormer
& $0.2476$
& $0.5825$
& $0.8192$
& $0.9395$
& $\mathbf{1.0761}$ \tabularnewline
F-FNO
& $\mathbf{0.0357}$
& $0.1021$
& $0.3118$
& $0.8598$
& $1.6042$ \tabularnewline
MemNO
& $0.0534$
& $0.1898$
& $0.4489$
& $0.8584$
& $1.3305$ \tabularnewline
PDE-Refiner
& $0.2638$
& $0.4556$
& $0.7727$
& $1.0484$
& $1.3378$ \tabularnewline
Poseidon
& $0.1387$
& $0.5443$
& $0.7654$
& $1.1764$
& $1.4233$ \tabularnewline
\textbf{CoMNO}
& $0.0486$
& $\mathbf{0.0938}$
& $\mathbf{0.2687}$
& $\mathbf{0.8506}$
& $1.1310$ \tabularnewline
\bottomrule
\end{tabular}
}
\ifdim\wd0>\linewidth
  \resizebox{\linewidth}{!}{\usebox0}%
\else
  \usebox0
\fi
\end{minipage}
\par
\noindent\begin{minipage}{\linewidth}
\centering
\footnotesize
\setlength{\tabcolsep}{3.2pt}
\renewcommand{\arraystretch}{1.13}
\captionof{table}{Rollout accuracy on Kolmogorov flow. Relative $\ell_2$ uses the first $H$ predictions; Final Step nRMSE uses step $197$.}
\label{tab:rollout_kolm}
\sbox0{%
\begin{tabular}{lccccc}
\toprule
Model
& \multicolumn{4}{c}{Rel. $\ell_2$ $\downarrow$}
& \shortstack{Final Step\\nRMSE} $\downarrow$ \tabularnewline
\cmidrule(lr){2-5}
\cmidrule(lr){6-6}
& $H=20$
& $H=50$
& $H=100$
& $H=197$
& $H=197$ \tabularnewline
\midrule
DPOT
& $0.4546$
& $0.7116$
& $0.8576$
& $0.9995$
& $1.1965$ \tabularnewline
FactFormer
& $0.4763$
& $0.6968$
& $0.8076$
& $\mathbf{0.8955}$
& $1.0062$ \tabularnewline
F-FNO
& $0.2020$
& $0.4352$
& $0.7602$
& $1.0692$
& $1.4734$ \tabularnewline
MemNO
& $0.3295$
& $0.6584$
& $0.8547$
& $1.0103$
& $1.2073$ \tabularnewline
PDE-Refiner
& $0.3691$
& $0.6988$
& $0.9162$
& $1.0740$
& $1.2081$ \tabularnewline
Poseidon
& $0.5038$
& $0.6749$
& $0.8693$
& $1.0546$
& $1.1934$ \tabularnewline
\textbf{CoMNO}
& $\mathbf{0.1243}$
& $\mathbf{0.2879}$
& $\mathbf{0.5689}$
& $0.9651$
& $\mathbf{0.9992}$ \tabularnewline
\bottomrule
\end{tabular}
}
\ifdim\wd0>\linewidth
  \resizebox{\linewidth}{!}{\usebox0}%
\else
  \usebox0
\fi
\end{minipage}
\par
\noindent\begin{minipage}{\linewidth}
\centering
\footnotesize
\setlength{\tabcolsep}{3.2pt}
\renewcommand{\arraystretch}{1.13}
\captionof{table}{Rollout accuracy on Cylinder. Relative $\ell_2$ uses the first $H$ predictions; Final Step nRMSE uses step $100$.}
\label{tab:rollout_cylinder}
\sbox0{%
\begin{tabular}{lcccc}
\toprule
Model
& \multicolumn{3}{c}{Rel. $\ell_2$ $\downarrow$}
& \shortstack{Final Step\\nRMSE} $\downarrow$ \tabularnewline
\cmidrule(lr){2-4}
\cmidrule(lr){5-5}
& $H=20$
& $H=50$
& $H=100$
& $H=100$ \tabularnewline
\midrule
DPOT
& $0.0738$
& $0.1010$
& $0.1353$
& $0.1896$ \tabularnewline
FactFormer
& $0.0746$
& $0.1105$
& $0.1500$
& $0.2047$ \tabularnewline
F-FNO
& $0.0658$
& $0.0955$
& $0.1274$
& $0.1707$ \tabularnewline
MemNO
& $0.0650$
& $0.0921$
& $0.1240$
& $0.1694$ \tabularnewline
PDE-Refiner
& $0.0613$
& $0.0909$
& $0.1254$
& $0.1752$ \tabularnewline
Poseidon
& $0.0734$
& $0.1001$
& $0.1314$
& $0.1769$ \tabularnewline
\textbf{CoMNO}
& $\mathbf{0.0588}$
& $\mathbf{0.0869}$
& $\mathbf{0.1188}$
& $\mathbf{0.1640}$ \tabularnewline
\bottomrule
\end{tabular}
}
\ifdim\wd0>\linewidth
  \resizebox{\linewidth}{!}{\usebox0}%
\else
  \usebox0
\fi
\end{minipage}
\par
\noindent\begin{minipage}{\linewidth}
\centering
\footnotesize
\setlength{\tabcolsep}{3.2pt}
\renewcommand{\arraystretch}{1.13}
\captionof{table}{Rollout accuracy on FSI. Relative $\ell_2$ uses the first $H$ predictions; Final Step nRMSE uses step $100$.}
\label{tab:rollout_fsi}
\sbox0{%
\begin{tabular}{lcccc}
\toprule
Model
& \multicolumn{3}{c}{Rel. $\ell_2$ $\downarrow$}
& \shortstack{Final Step\\nRMSE} $\downarrow$ \tabularnewline
\cmidrule(lr){2-4}
\cmidrule(lr){5-5}
& $H=20$
& $H=50$
& $H=100$
& $H=100$ \tabularnewline
\midrule
DPOT
& $0.1021$
& $0.1550$
& $0.2425$
& $0.4233$ \tabularnewline
FactFormer
& $0.0707$
& $0.1260$
& $0.1970$
& $0.2943$ \tabularnewline
F-FNO
& $0.0690$
& $0.1227$
& $0.1967$
& $0.3021$ \tabularnewline
MemNO
& $0.0631$
& $0.1091$
& $0.1645$
& $0.2397$ \tabularnewline
PDE-Refiner
& $0.0676$
& $0.1195$
& $0.1852$
& $0.2786$ \tabularnewline
Poseidon
& $0.1035$
& $0.1593$
& $0.2467$
& $0.4357$ \tabularnewline
\textbf{CoMNO}
& $\mathbf{0.0610}$
& $\mathbf{0.1080}$
& $\mathbf{0.1607}$
& $\mathbf{0.2318}$ \tabularnewline
\bottomrule
\end{tabular}
}
\ifdim\wd0>\linewidth
  \resizebox{\linewidth}{!}{\usebox0}%
\else
  \usebox0
\fi
\end{minipage}
\par
\noindent\begin{minipage}{\linewidth}
\centering
\footnotesize
\setlength{\tabcolsep}{3.2pt}
\renewcommand{\arraystretch}{1.13}
\captionof{table}{Full-rollout diagnostics on Burgers ($197$ steps). Late Step nRMSE uses the final half of the forecast; Energy Curve Error measures relative error in energy evolution.}
\label{tab:diagnostic_burgers}
\sbox0{%
\begin{tabular}{lccccc}
\toprule
Model
& \shortstack{Trajectory\\nRMSE} $\downarrow$
& \shortstack{Final Step\\nRMSE} $\downarrow$
& \shortstack{Late Step\\nRMSE} $\downarrow$
& Mean Corr. $\uparrow$
& \shortstack{Energy Curve\\Error} $\downarrow$ \tabularnewline
\midrule
DPOT
& $0.7323$
& $0.9955$
& $0.9867$
& $0.6244$
& $0.6728$ \tabularnewline
FactFormer
& $0.1830$
& $0.3433$
& $0.2786$
& $0.9854$
& $0.4130$ \tabularnewline
F-FNO
& $1.5670$
& $3.6467$
& $3.0536$
& $0.5999$
& $11.6846$ \tabularnewline
MemNO
& $0.4121$
& $0.7902$
& $0.6757$
& $0.9158$
& $0.7061$ \tabularnewline
PDE-Refiner
& $0.8474$
& $1.7102$
& $1.5007$
& $0.6686$
& $1.7539$ \tabularnewline
Poseidon
& $0.7684$
& $1.2365$
& $1.3627$
& $0.6837$
& $0.6421$ \tabularnewline
\textbf{CoMNO}
& $\mathbf{0.1805}$
& $\mathbf{0.2812}$
& $\mathbf{0.2657}$
& $\mathbf{0.9868}$
& $\mathbf{0.3924}$ \tabularnewline
\bottomrule
\end{tabular}
}
\ifdim\wd0>\linewidth
  \resizebox{\linewidth}{!}{\usebox0}%
\else
  \usebox0
\fi
\end{minipage}
\par
\noindent\begin{minipage}{\linewidth}
\centering
\footnotesize
\setlength{\tabcolsep}{3.2pt}
\renewcommand{\arraystretch}{1.13}
\captionof{table}{Full-rollout diagnostics on KS ($197$ steps). Late Step nRMSE uses the final half of the forecast; Energy Curve Error measures relative error in energy evolution.}
\label{tab:diagnostic_ks}
\sbox0{%
\begin{tabular}{lccccc}
\toprule
Model
& \shortstack{Trajectory\\nRMSE} $\downarrow$
& \shortstack{Final Step\\nRMSE} $\downarrow$
& \shortstack{Late Step\\nRMSE} $\downarrow$
& Mean Corr. $\uparrow$
& \shortstack{Energy Curve\\Error} $\downarrow$ \tabularnewline
\midrule
DPOT
& $1.0595$
& $1.3979$
& $1.3777$
& $0.3390$
& $0.1303$ \tabularnewline
FactFormer
& $0.9011$
& $\mathbf{1.0761}$
& $\mathbf{1.0547}$
& $0.3625$
& $0.7322$ \tabularnewline
F-FNO
& $0.6890$
& $1.6042$
& $1.3032$
& $\mathbf{0.6872}$
& $0.9159$ \tabularnewline
MemNO
& $0.7438$
& $1.3305$
& $1.2119$
& $0.6533$
& $0.1908$ \tabularnewline
PDE-Refiner
& $0.9837$
& $1.3378$
& $1.2963$
& $0.4393$
& $0.1233$ \tabularnewline
Poseidon
& $1.0235$
& $1.4233$
& $1.2553$
& $0.4297$
& $0.1684$ \tabularnewline
\textbf{CoMNO}
& $\mathbf{0.6630}$
& $1.1310$
& $1.0957$
& $0.6732$
& $\mathbf{0.0941}$ \tabularnewline
\bottomrule
\end{tabular}
}
\ifdim\wd0>\linewidth
  \resizebox{\linewidth}{!}{\usebox0}%
\else
  \usebox0
\fi
\end{minipage}
\par
\noindent\begin{minipage}{\linewidth}
\centering
\footnotesize
\setlength{\tabcolsep}{3.2pt}
\renewcommand{\arraystretch}{1.13}
\captionof{table}{Full-rollout diagnostics on Kolmogorov flow ($197$ steps). Late Step nRMSE uses the final half of the forecast; Energy Curve Error measures relative error in energy evolution.}
\label{tab:diagnostic_kolm}
\sbox0{%
\begin{tabular}{lccccc}
\toprule
Model
& \shortstack{Trajectory\\nRMSE} $\downarrow$
& \shortstack{Final Step\\nRMSE} $\downarrow$
& \shortstack{Late Step\\nRMSE} $\downarrow$
& Mean Corr. $\uparrow$
& \shortstack{Energy Curve\\Error} $\downarrow$ \tabularnewline
\midrule
DPOT
& $0.9768$
& $1.1965$
& $1.1565$
& $0.3991$
& $0.3246$ \tabularnewline
FactFormer
& $0.9816$
& $1.0062$
& $\mathbf{0.9933}$
& $0.5200$
& $0.3585$ \tabularnewline
F-FNO
& $0.9966$
& $1.4734$
& $1.3657$
& $0.5210$
& $0.7592$ \tabularnewline
MemNO
& $0.9773$
& $1.2073$
& $1.1768$
& $0.4079$
& $0.3163$ \tabularnewline
PDE-Refiner
& $1.0418$
& $1.2081$
& $1.2322$
& $0.2517$
& $0.4942$ \tabularnewline
Poseidon
& $0.9822$
& $1.1934$
& $1.1652$
& $0.4228$
& $0.3657$ \tabularnewline
\textbf{CoMNO}
& $\mathbf{0.7886}$
& $\mathbf{0.9992}$
& $1.0294$
& $\mathbf{0.6222}$
& $\mathbf{0.2164}$ \tabularnewline
\bottomrule
\end{tabular}
}
\ifdim\wd0>\linewidth
  \resizebox{\linewidth}{!}{\usebox0}%
\else
  \usebox0
\fi
\end{minipage}
\par
\noindent\begin{minipage}{\linewidth}
\centering
\footnotesize
\setlength{\tabcolsep}{3.2pt}
\renewcommand{\arraystretch}{1.13}
\captionof{table}{Full-rollout diagnostics on Cylinder ($100$ steps). Late Step nRMSE uses the final half of the forecast; Energy Curve Error measures relative error in energy evolution.}
\label{tab:diagnostic_cylinder}
\sbox0{%
\begin{tabular}{lccccc}
\toprule
Model
& \shortstack{Trajectory\\nRMSE} $\downarrow$
& \shortstack{Final Step\\nRMSE} $\downarrow$
& \shortstack{Late Step\\nRMSE} $\downarrow$
& Mean Corr. $\uparrow$
& \shortstack{Energy Curve\\Error} $\downarrow$ \tabularnewline
\midrule
DPOT
& $0.1292$
& $0.1896$
& $0.1693$
& $0.9840$
& $0.2070$ \tabularnewline
FactFormer
& $0.1422$
& $0.2047$
& $0.1883$
& $0.9831$
& $0.2099$ \tabularnewline
F-FNO
& $0.1216$
& $0.1707$
& $0.1587$
& $0.9871$
& $0.2249$ \tabularnewline
MemNO
& $0.1181$
& $0.1694$
& $0.1554$
& $0.9877$
& $0.2261$ \tabularnewline
PDE-Refiner
& $0.1189$
& $0.1752$
& $0.1590$
& $0.9873$
& $0.0807$ \tabularnewline
Poseidon
& $0.1267$
& $0.1769$
& $0.1687$
& $0.9826$
& $0.2081$ \tabularnewline
\textbf{CoMNO}
& $\mathbf{0.1127}$
& $\mathbf{0.1640}$
& $\mathbf{0.1503}$
& $\mathbf{0.9888}$
& $\mathbf{0.0497}$ \tabularnewline
\bottomrule
\end{tabular}
}
\ifdim\wd0>\linewidth
  \resizebox{\linewidth}{!}{\usebox0}%
\else
  \usebox0
\fi
\end{minipage}
\par
\noindent\begin{minipage}{\linewidth}
\centering
\footnotesize
\setlength{\tabcolsep}{3.2pt}
\renewcommand{\arraystretch}{1.13}
\captionof{table}{Full-rollout diagnostics on FSI ($100$ steps). Late Step nRMSE uses the final half of the forecast; Energy Curve Error measures relative error in energy evolution.}
\label{tab:diagnostic_fsi}
\sbox0{%
\begin{tabular}{lccccc}
\toprule
Model
& \shortstack{Trajectory\\nRMSE} $\downarrow$
& \shortstack{Final Step\\nRMSE} $\downarrow$
& \shortstack{Late Step\\nRMSE} $\downarrow$
& Mean Corr. $\uparrow$
& \shortstack{Energy Curve\\Error} $\downarrow$ \tabularnewline
\midrule
DPOT
& $0.2241$
& $0.4233$
& $0.3238$
& $0.9628$
& $0.0754$ \tabularnewline
FactFormer
& $0.1818$
& $0.2943$
& $0.2636$
& $0.9752$
& $\mathbf{0.0301}$ \tabularnewline
F-FNO
& $0.1805$
& $0.3021$
& $0.2662$
& $0.9753$
& $0.0328$ \tabularnewline
MemNO
& $0.1530$
& $0.2397$
& $0.2169$
& $0.9828$
& $0.0519$ \tabularnewline
PDE-Refiner
& $0.1712$
& $0.2786$
& $0.2474$
& $0.9786$
& $0.0398$ \tabularnewline
Poseidon
& $0.2457$
& $0.4357$
& $0.3485$
& $0.9583$
& $0.0867$ \tabularnewline
\textbf{CoMNO}
& $\mathbf{0.1493}$
& $\mathbf{0.2318}$
& $\mathbf{0.2105}$
& $\mathbf{0.9833}$
& $0.0314$ \tabularnewline
\bottomrule
\end{tabular}
}
\ifdim\wd0>\linewidth
  \resizebox{\linewidth}{!}{\usebox0}%
\else
  \usebox0
\fi
\end{minipage}
\par

\subsection{Rollout Stability}
\label{sec:rollout_supp}

\paragraph{Recursive evaluation.}
We extend prediction beyond the supervised 20-step window
using the four observed input frames specified in the main text.
APEBench rollouts continue autoregressively for 197 steps;
RealPDEBench rollouts reach 100 steps by recursively chaining
five 20-step forecast blocks. Predicted states provide subsequent
inputs, and CoMNO carries its recurrent memory along the rollout.
Tables~\ref{tab:rollout_burgers} to~\ref{tab:rollout_fsi}
report relative $\ell_2$ error over the first $H$ predicted steps
and Final Step nRMSE at the longest evaluated horizon.
The 100-step APEBench plots display prefixes of these extended
rollouts. All metrics use inverse-normalized fields and the
channel conventions defined in the main text. Entries report means;
boldface marks the best mean in each column.

\paragraph{Complementary error measures.}
Prefix relative $\ell_2$ aggregates squared errors and reference
magnitudes over time, channels, and space before taking their
norm ratio. Trajectory nRMSE averages the stepwise errors
$e_t$ from Eq.~\eqref{eq:stepwise_nrmse} with equal temporal
weight, while Final Step nRMSE measures the last prediction.
These summaries capture different aspects of rollout accuracy
and can produce different rankings.
Tables~\ref{tab:diagnostic_burgers} to~\ref{tab:diagnostic_fsi}
report diagnostics over the full 197-step APEBench or 100-step
RealPDEBench rollout. Their terminal entries are identical to
those in the corresponding horizon tables.
Late Step nRMSE averages $e_t$ over the final half of the forecast;
the final-quarter metric in the commutator intervention remains
a separate diagnostic. Mean Corr.\ follows the main-text
stepwise Pearson correlation. Energy Curve Error averages the
absolute predicted--reference energy difference divided by the
reference energy plus $10^{-6}$. Energy is the spatial mean
of the sum of squared evaluated field components.

\paragraph{Accuracy across horizons.}
CoMNO achieves the lowest prefix relative $\ell_2$ at every
reported horizon on Burgers, Cylinder, and FSI, together with
the lowest terminal nRMSE. The Burgers and Cylinder results
agree with Figure~\ref{fig:rollout}.
On KS, F-FNO leads at 20 steps, whereas CoMNO leads at
50, 100, and 197 steps; FactFormer achieves the lowest
terminal nRMSE. On Kolmogorov flow, CoMNO leads at
20, 50, and 100 steps and has the lowest terminal nRMSE,
while FactFormer achieves the lowest 197-step prefix error.
The horizon and terminal results therefore distinguish
accuracy accumulated over the forecast from accuracy at
its endpoint.

\paragraph{Late-rollout and physical fidelity.}
CoMNO achieves the lowest Trajectory nRMSE on all five
systems. It also leads Late Step nRMSE on Burgers, Cylinder,
and FSI, while FactFormer leads this metric on KS and
Kolmogorov flow. CoMNO has the highest Mean Corr.\ on
four systems, with F-FNO leading on KS. Energy Curve Error
is lowest for CoMNO on Burgers, KS, Kolmogorov flow, and
Cylinder, and for FactFormer on FSI.
Together, these diagnostics show that CoMNO's trajectory
accuracy is accompanied by strong late-rollout performance
and faithful energy evolution across diverse dynamics.

\par
\noindent
\begin{minipage}[t]{0.485\linewidth}
\centering
\footnotesize
\setlength{\tabcolsep}{3.2pt}
\renewcommand{\arraystretch}{1.13}

\captionof{table}{
Physical diagnostics on Burgers. StepRes measures normalized one-step discrepancy from reference numerical evolution (Eq.~\eqref{eq:physical_field_metrics}); the remaining metric measures error relative to the reference evolution.
}
\label{tab:physics_diagnostics_burgers}

\vspace{0.4em}

\resizebox{\linewidth}{!}{%
\begin{tabular}{lcc}
\toprule
Model
& StepRes $\downarrow$
& \shortstack{Energy Curve\\Rel. Error} $\downarrow$ \tabularnewline
\midrule
DPOT
& $0.1943$
& $0.1459$ \tabularnewline
FactFormer
& $0.0410$
& $0.0842$ \tabularnewline
F-FNO
& $0.9517$
& $2.0950$ \tabularnewline
MemNO
& $0.0428$
& $0.1637$ \tabularnewline
PDE-Refiner
& $0.7718$
& $0.3525$ \tabularnewline
Poseidon
& $0.2437$
& $0.1956$ \tabularnewline
\textbf{CoMNO}
& $\mathbf{0.0354}$
& $\mathbf{0.0751}$ \tabularnewline
\bottomrule
\end{tabular}%
}
\end{minipage}
\hfill
\begin{minipage}[t]{0.485\linewidth}
\centering
\footnotesize
\setlength{\tabcolsep}{3.2pt}
\renewcommand{\arraystretch}{1.13}

\captionof{table}{
Physical diagnostics on KS. StepRes measures normalized one-step discrepancy from reference numerical evolution (Eq.~\eqref{eq:physical_field_metrics}); the remaining metric measures error relative to the reference evolution.
}
\label{tab:physics_diagnostics_ks}

\vspace{0.4em}

\resizebox{\linewidth}{!}{%
\begin{tabular}{lcc}
\toprule
Model
& StepRes $\downarrow$
& \shortstack{Gradient-Energy Curve\\Rel. Error} $\downarrow$ \tabularnewline
\midrule
DPOT
& $0.3730$
& $1.1010$ \tabularnewline
FactFormer
& $0.0501$
& $0.7391$ \tabularnewline
F-FNO
& $0.2592$
& $0.2457$ \tabularnewline
MemNO
& $0.0972$
& $0.2882$ \tabularnewline
PDE-Refiner
& $0.5851$
& $0.3383$ \tabularnewline
Poseidon
& $0.3856$
& $2.4023$ \tabularnewline
\textbf{CoMNO}
& $\mathbf{0.0091}$
& $\mathbf{0.1135}$ \tabularnewline
\bottomrule
\end{tabular}%
}
\end{minipage}
\par

\noindent
\begin{minipage}[t]{0.485\linewidth}
\centering
\footnotesize
\setlength{\tabcolsep}{3.2pt}
\renewcommand{\arraystretch}{1.13}

\captionof{table}{
Physical diagnostics on Cylinder. Vorticity error measures spatial rotational-structure error; enstrophy error measures aggregate rotational-intensity discrepancy.
}
\label{tab:physics_diagnostics_cylinder}

\vspace{0.4em}

\resizebox{\linewidth}{!}{%
\begin{tabular}{lcc}
\toprule
Model
& \shortstack{Vorticity Rel.\\$\ell_2$} $\downarrow$
& \shortstack{Enstrophy Rel.\\Error} $\downarrow$ \tabularnewline
\midrule
DPOT
& $0.6939$
& $0.5250$ \tabularnewline
FactFormer
& $0.6920$
& $0.3945$ \tabularnewline
F-FNO
& $0.6125$
& $0.2594$ \tabularnewline
MemNO
& $0.6109$
& $0.3747$ \tabularnewline
PDE-Refiner
& $0.6085$
& $0.3080$ \tabularnewline
Poseidon
& $0.6727$
& $0.3977$ \tabularnewline
\textbf{CoMNO}
& $\mathbf{0.5979}$
& $\mathbf{0.2508}$ \tabularnewline
\bottomrule
\end{tabular}%
}
\end{minipage}
\hfill
\begin{minipage}[t]{0.485\linewidth}
\centering
\footnotesize
\setlength{\tabcolsep}{3.2pt}
\renewcommand{\arraystretch}{1.13}

\captionof{table}{
Physical diagnostics on FSI. Vorticity error measures spatial rotational-structure error; enstrophy error measures aggregate rotational-intensity discrepancy.
}
\label{tab:physics_diagnostics_fsi}

\vspace{0.4em}

\resizebox{\linewidth}{!}{%
\begin{tabular}{lcc}
\toprule
Model
& \shortstack{Vorticity Rel.\\$\ell_2$} $\downarrow$
& \shortstack{Enstrophy Rel.\\Error} $\downarrow$ \tabularnewline
\midrule
DPOT
& $0.6835$
& $0.2028$ \tabularnewline
FactFormer
& $0.5996$
& $0.2648$ \tabularnewline
F-FNO
& $0.5622$
& $0.2492$ \tabularnewline
MemNO
& $0.5126$
& $0.2317$ \tabularnewline
PDE-Refiner
& $0.5550$
& $0.2320$ \tabularnewline
Poseidon
& $0.7246$
& $0.2137$ \tabularnewline
\textbf{CoMNO}
& $\mathbf{0.5109}$
& $\mathbf{0.2010}$ \tabularnewline
\bottomrule
\end{tabular}%
}
\end{minipage}
\par

\noindent
\begin{minipage}{\linewidth}
\centering
\footnotesize
\setlength{\tabcolsep}{5pt}
\renewcommand{\arraystretch}{1.13}

\captionof{table}{
Physical diagnostics on Kolmogorov flow. StepRes measures normalized one-step discrepancy from reference numerical evolution (Eq.~\eqref{eq:physical_field_metrics}); KE and enstrophy errors measure deviations from the reference evolution.
}
\label{tab:physics_diagnostics_kolm}

\vspace{0.4em}

\begin{tabular}{lccc}
\toprule
Model
& StepRes $\downarrow$
& KE Error $\downarrow$
& Enstrophy Error $\downarrow$ \tabularnewline
\midrule
DPOT
& $0.395$
& $0.346$
& $0.396$ \tabularnewline
FactFormer
& $0.155$
& $0.304$
& $0.387$ \tabularnewline
F-FNO
& $0.655$
& $5.618$
& $0.420$ \tabularnewline
MemNO
& $0.151$
& $0.234$
& $\mathbf{0.306}$ \tabularnewline
PDE-Refiner
& $0.380$
& $0.645$
& $0.509$ \tabularnewline
Poseidon
& $0.304$
& $0.561$
& $0.566$ \tabularnewline
\textbf{CoMNO}
& $\mathbf{0.136}$
& $\mathbf{0.175}$
& $0.368$ \tabularnewline
\bottomrule
\end{tabular}

\end{minipage}
\par

\subsection{Physical Consistency}
\label{sec:physics_supp}

\paragraph{Diagnostics.}
StepRes and VortRel follow Eq.~\eqref{eq:physical_field_metrics}.
StepRes measures the relative discrepancy between each next
prediction and the APEBench reference time-stepper applied to
the preceding prediction. KS and Kolmogorov flow use their
corresponding reference time-steppers.
VortRel measures relative error in the spatial vorticity field.
Energy, gradient-energy, and kinetic-energy errors divide the
temporal RMS of the predicted-reference observable difference
by the reference curve's RMS, floored at $\epsilon_{\mathrm{phys}}$.
Enstrophy error averages stepwise relative absolute differences.
Gradient energy is the spatial mean of squared field gradients;
enstrophy is the spatial mean of squared vorticity.

\paragraph{Result.}
Tables~\ref{tab:physics_diagnostics_burgers}--\ref{tab:physics_diagnostics_kolm}
report the physical diagnostics across all systems. CoMNO combines
the lowest StepRes with the closest energy evolution on Burgers
and gradient-energy evolution on KS. On Kolmogorov flow it leads
StepRes and kinetic-energy accuracy; MemNO has the lowest
enstrophy error. On Cylinder and FSI, CoMNO leads both vorticity
and enstrophy errors. Together, these improvements connect lower
field-prediction error to more accurate spatial vorticity and
aggregate rotational intensity.

\par
\noindent\begin{minipage}{\linewidth}
\centering
\footnotesize
\setlength{\tabcolsep}{3.2pt}
\renewcommand{\arraystretch}{1.13}
\captionof{table}{Ablation on Burgers. Each variant removes one of the three modules evaluated in the main text.}
\label{tab:ablation_burgers}
\sbox0{%
\begin{tabular}{>{\raggedright\arraybackslash}p{.18\linewidth}ccccc}
\toprule
Model & Rel. $\ell_2$ $\downarrow$ & Traj. nRMSE $\downarrow$ & Final nRMSE $\downarrow$ & HF Error $\downarrow$ & Mean Corr. $\uparrow$ \\
\midrule
w/o Operator Commutator & $0.0738$ & $0.0783$ & $0.1848$ & $1.0506$ & $0.9966$ \\
w/o Closure Memory & $0.0716$ & $0.0716$ & $0.1946$ & $0.5029$ & $0.9958$ \\
w/o Closure Experts & $0.0650$ & $0.0650$ & $0.1775$ & $0.5624$ & $0.9952$ \\
\midrule
\textbf{CoMNO} & $\mathbf{0.0568}$ & $\mathbf{0.0568}$ & $\mathbf{0.1412}$ & $\mathbf{0.4688}$ & $\mathbf{0.9978}$ \\
\bottomrule
\end{tabular}
}
\ifdim\wd0>\linewidth
  \resizebox{\linewidth}{!}{\usebox0}%
\else
  \usebox0
\fi
\end{minipage}
\par
\noindent\begin{minipage}{\linewidth}
\centering
\footnotesize
\setlength{\tabcolsep}{3.2pt}
\renewcommand{\arraystretch}{1.13}
\captionof{table}{Ablation on Cylinder. Each variant removes one of the three modules evaluated in the main text.}
\label{tab:ablation_cylinder}
\sbox0{%
\begin{tabular}{>{\raggedright\arraybackslash}p{.18\linewidth}ccccc}
\toprule
Model & Rel. $\ell_2$ $\downarrow$ & Traj. nRMSE $\downarrow$ & Final nRMSE $\downarrow$ & HF Error $\downarrow$ & Mean Corr. $\uparrow$ \\
\midrule
w/o Operator Commutator & $0.1328$ & $0.1252$ & $0.1911$ & $0.8212$ & $0.9858$ \\
w/o Closure Memory & $0.1405$ & $0.1325$ & $0.2034$ & $0.8263$ & $0.9841$ \\
w/o Closure Experts & $0.1269$ & $0.1210$ & $0.1810$ & $0.8113$ & $0.9872$ \\
\midrule
\textbf{CoMNO} & $\mathbf{0.1181}$ & $\mathbf{0.1181}$ & $\mathbf{0.1718}$ & $\mathbf{0.8080}$ & $\mathbf{0.9874}$ \\
\bottomrule
\end{tabular}
}
\ifdim\wd0>\linewidth
  \resizebox{\linewidth}{!}{\usebox0}%
\else
  \usebox0
\fi
\end{minipage}
\par

\subsection{Ablation Study}
\label{sec:ablation_supp}

\paragraph{Module removals.}
We retrain the three variants in the main text under the same
data splits, normalization, training windows, optimization,
and checkpoint-selection rule. Removing
\emph{Operator Commutator} disables cross-scale evidence and its
state and memory discrepancy penalties. Removing \emph{Closure Memory}
disables the recurrent closure state and its coupling to local features.
Removing \emph{Closure Experts} disables the adaptive closure
correction and its magnitude, diversity, and routing-entropy penalties.
Tables~\ref{tab:ablation_burgers} and~\ref{tab:ablation_cylinder}
compare these variants with the full CoMNO model using the five
tabulated main-text prediction metrics.

\paragraph{Contributions to prediction.}
The full model leads all five metrics on both systems. On Burgers,
removing Operator Commutator causes the largest increase in relative
$\ell_2$, trajectory nRMSE, and HF Error. Removing Closure Memory
causes the largest terminal-error increase, while removing Closure
Experts produces the largest loss of spatial correlation. The
three components therefore contribute differently to trajectory,
endpoint, and spatial accuracy. On Cylinder, removing Closure Memory
causes the largest degradation across all five metrics, making
recurrent context particularly important for this measured-flow task.
Removing either Operator Commutator or Closure Experts also worsens
every metric, supporting their combined use with Closure Memory.

\par
\noindent\begin{minipage}{\linewidth}
\centering

\begin{minipage}[t]{.485\linewidth}
\centering
\footnotesize
\setlength{\tabcolsep}{3pt}
\renewcommand{\arraystretch}{1.08}

\captionof{table}{Timing on Burgers.}
\label{tab:efficiency_burgers}

\begin{tabular}{@{}lcc@{}}
\toprule
Model
& \shortstack{Training\\s/run $\downarrow$}
& \shortstack{Inference\\ms/window $\downarrow$} \tabularnewline
\midrule
DPOT
& $3025.5905$
& $1650.1274$ \tabularnewline
FactFormer
& $59327.3033$
& $1322.1693$ \tabularnewline
F-FNO
& $\mathbf{1856.1637}$
& $884.5422$ \tabularnewline
MemNO
& $54294.9713$
& $7606.0083$ \tabularnewline
PDE-Refiner
& $3397.8745$
& $1636.9655$ \tabularnewline
Poseidon
& $3123.3230$
& $\mathbf{735.9401}$ \tabularnewline
\textbf{CoMNO}
& $10166.1472$
& $770.1263$ \tabularnewline
\bottomrule
\end{tabular}

\end{minipage}
\hfill
\begin{minipage}[t]{.485\linewidth}
\centering
\footnotesize
\setlength{\tabcolsep}{3pt}
\renewcommand{\arraystretch}{1.08}

\captionof{table}{Timing on KS.}
\label{tab:efficiency_ks}

\begin{tabular}{@{}lcc@{}}
\toprule
Model
& \shortstack{Training\\s/run $\downarrow$}
& \shortstack{Inference\\ms/window $\downarrow$} \tabularnewline
\midrule
DPOT
& $10492.0486$
& $1835.1998$ \tabularnewline
FactFormer
& $44211.3145$
& $1331.7986$ \tabularnewline
F-FNO
& $\mathbf{3695.9106}$
& $809.9820$ \tabularnewline
MemNO
& $64371.3925$
& $4562.4230$ \tabularnewline
PDE-Refiner
& $24183.0945$
& $1740.7213$ \tabularnewline
Poseidon
& $10604.6931$
& $\mathbf{587.7483}$ \tabularnewline
\textbf{CoMNO}
& $15090.1923$
& $918.9086$ \tabularnewline
\bottomrule
\end{tabular}

\end{minipage}

\par

\begin{minipage}[t]{.485\linewidth}
\centering
\footnotesize
\setlength{\tabcolsep}{3pt}
\renewcommand{\arraystretch}{1.08}

\captionof{table}{Timing on Kolmogorov flow.}
\label{tab:efficiency_kolm}

\begin{tabular}{@{}lcc@{}}
\toprule
Model
& \shortstack{Training\\s/run $\downarrow$}
& \shortstack{Inference\\ms/window $\downarrow$} \tabularnewline
\midrule
DPOT
& $\mathbf{3619.6117}$
& $1857.3645$ \tabularnewline
FactFormer
& $14742.0839$
& $1344.8393$ \tabularnewline
F-FNO
& $6178.5432$
& $759.3715$ \tabularnewline
MemNO
& $67652.6916$
& $1839.8797$ \tabularnewline
PDE-Refiner
& $8849.2068$
& $1698.4824$ \tabularnewline
Poseidon
& $5008.0558$
& $\mathbf{533.5882}$ \tabularnewline
\textbf{CoMNO}
& $13160.5317$
& $623.1423$ \tabularnewline
\bottomrule
\end{tabular}

\end{minipage}
\hfill
\begin{minipage}[t]{.485\linewidth}
\centering
\footnotesize
\setlength{\tabcolsep}{3pt}
\renewcommand{\arraystretch}{1.08}

\captionof{table}{Timing on Cylinder.}
\label{tab:efficiency_cylinder}

\begin{tabular}{@{}lcc@{}}
\toprule
Model
& \shortstack{Training\\s/run $\downarrow$}
& \shortstack{Inference\\ms/window $\downarrow$} \tabularnewline
\midrule
DPOT
& $20071.4915$
& $256.9994$ \tabularnewline
FactFormer
& $129743.4437$
& $633.8724$ \tabularnewline
F-FNO
& $\mathbf{16556.6802}$
& $397.9526$ \tabularnewline
MemNO
& $282159.0751$
& $750.1990$ \tabularnewline
PDE-Refiner
& $32084.0626$
& $571.6319$ \tabularnewline
Poseidon
& $18560.8863$
& $\mathbf{167.8150}$ \tabularnewline
\textbf{CoMNO}
& $98906.4574$
& $937.2157$ \tabularnewline
\bottomrule
\end{tabular}

\end{minipage}

\par

\begin{minipage}[t]{\linewidth}
\centering
\footnotesize
\setlength{\tabcolsep}{6pt}
\renewcommand{\arraystretch}{1.08}

\captionof{table}{Timing on FSI.}
\label{tab:efficiency_fsi}

\begin{tabular}{@{}lcc@{}}
\toprule
Model
& \shortstack{Training\\s/run $\downarrow$}
& \shortstack{Inference\\ms/window $\downarrow$} \tabularnewline
\midrule
DPOT
& $19286.0141$
& $\mathbf{141.7249}$ \tabularnewline
FactFormer
& $157273.1993$
& $893.1758$ \tabularnewline
F-FNO
& $\mathbf{15575.2524}$
& $142.2549$ \tabularnewline
MemNO
& $272467.3243$
& $1357.5088$ \tabularnewline
PDE-Refiner
& $26305.7649$
& $1342.5798$ \tabularnewline
Poseidon
& $18151.8495$
& $481.2714$ \tabularnewline
\textbf{CoMNO}
& $124610.9605$
& $1672.3246$ \tabularnewline
\bottomrule
\end{tabular}

\end{minipage}

\end{minipage}
\par

\section{Hyperparameters and Computational Cost}
\label{app:implementation_cost}

\paragraph{Model configuration.}
The local encoder and base operator use width $64$, with two
local blocks and four base blocks. Closure Memory has width $32$ and
operates on a grid downsampled by a factor of two. The base
operator retains $12$ Fourier modes and applies a spectral block
every two blocks. Operator Commutator uses scales $\{2,4\}$ and
is recomputed every two prediction steps; intermediate steps use
zero current evidence. Gradients propagate through this construction.
Closure Experts comprise five expert families, with the boundary expert active
on non-periodic domains and circular padding on periodic domains.

\paragraph{Optimization.}
CoMNO uses AdamW with a constant learning rate of $2\times10^{-4}$,
weight decay $10^{-4}$, batch size $4$, and gradient clipping at
$1.0$. Four observed frames initialize a $20$-step training rollout,
with teacher forcing disabled. Training runs for at most $500$
epochs, with validation after each epoch and early stopping after
$10$ epochs without improvement. All methods use the same
data splits, forecast windows, and metric implementation.
Normalization statistics are fitted to each modality's training split.

\paragraph{Loss weights.}
The rollout loss has unit weight and includes a trajectory-relative
term weighted by $0.05$. Temporal weights have unit mean and increase
linearly, with the final-step weight twice the first-step weight.
The Fourier-amplitude loss has weight $5\times10^{-3}$. Raw state
and memory commutator penalties have weights $0.02$ and $0.01$.
Memory and closure-magnitude regularization each use $10^{-4}$;
expert-diversity and routing-entropy terms use $10^{-3}$ and
$10^{-5}$. The temporal component of memory regularization has
coefficient $0.1$.

\paragraph{Validation selection.}
Checkpoints minimize a validation score consisting of full-trajectory
relative $\ell_2$ plus half the late-window nRMSE and half the final-step
nRMSE. The late window is the final third of the forecast.
Architecture and loss weights are selected on Burgers validation
data and transferred across systems, with physical channels,
grid spacing, and boundary handling set for each task. Baselines
retain their official architectures and pretrained initializations
where applicable, while following the common training and evaluation
protocol stated in the main text, including the shared epoch budget,
validation schedule, and checkpoint criterion.

\paragraph{Computational cost.}
Tables~\ref{tab:efficiency_burgers}--\ref{tab:efficiency_fsi}
report training time per run and inference time per complete
forecast window. Training includes optimization, validation,
and checkpoint saving. Inference includes memory initialization,
recursive prediction, commutator or refinement computations,
denormalization, and device synchronization, with batch size one
and gradients disabled. Runs use one NVIDIA RTX PRO 6000 GPU
with 96\,GB memory, PyTorch 2.8.0, and CUDA 12.8.

CoMNO combines competitive APEBench inference time with lower
prediction errors. Its inference cost is higher on Cylinder
and FSI. Each commutator update adds a provisional transition
on the observation grid and each configured coarse grid, so
the timing results account for the computation used to obtain
the accuracy gains.

\end{document}